\documentclass{article}

\usepackage{microtype}
\usepackage{graphicx}
\usepackage{subcaption}
\usepackage{booktabs} 

\usepackage[backref=page]{hyperref}

\usepackage[preprint]{icml2026}

\usepackage{amsmath}
\usepackage{amssymb}
\usepackage{mathtools}
\usepackage{amsthm}

\usepackage[capitalize,noabbrev]{cleveref}

\theoremstyle{plain}

\theoremstyle{definition}

\theoremstyle{remark}

\usepackage[textsize=tiny]{todonotes}

\usepackage{xcolor}
\definecolor{myred}{RGB}{200,0,16}
\definecolor{mygreen}{RGB}{27, 142, 66}
\definecolor{myblue}{RGB}{15, 103, 202}

\definecolor{1st}{RGB}{193, 86, 94}
\definecolor{2nd}{RGB}{130, 173, 127}
\definecolor{3rd}{RGB}{126, 164, 209}
\definecolor{top1p}{RGB}{183, 183, 235}
\definecolor{totalp}{RGB}{234, 184, 131}

\usepackage{multirow}
\usepackage{colortbl}

\icmltitlerunning{\textit{arXiv preprint}}

\begin{document}

\twocolumn[{
  \icmltitle{RealCamo: Boosting Real Camouflage Synthesis with Layout Controls and Textual-Visual Guidance}

  \begin{icmlauthorlist}
    \icmlauthor{Chunyuan Chen}{1}
    \icmlauthor{Yunuo Cai}{2}
    \icmlauthor{Shujuan Li}{1}
    \icmlauthor{Weiyun Liang}{1}
    \icmlauthor{Bin Wang}{1}
    \icmlauthor{Jing Xu}{1}
  \end{icmlauthorlist}

  \icmlaffiliation{1}{College of Artificial Intelligence, Nankai University, Tianjin, China}
  \icmlaffiliation{2}{School of Data Science, Fudan University, Shanghai, China}

  \icmlcorrespondingauthor{Jing Xu}{xujing@nankai.edu.cn}
  
  \icmlkeywords{Camouflaged Image Generation, Diffusion Model, Controlled Image Generation, Synthetic Data}
  
  \vskip 0.3in
  
  \begin{center}
    \includegraphics[width=1.0\textwidth]{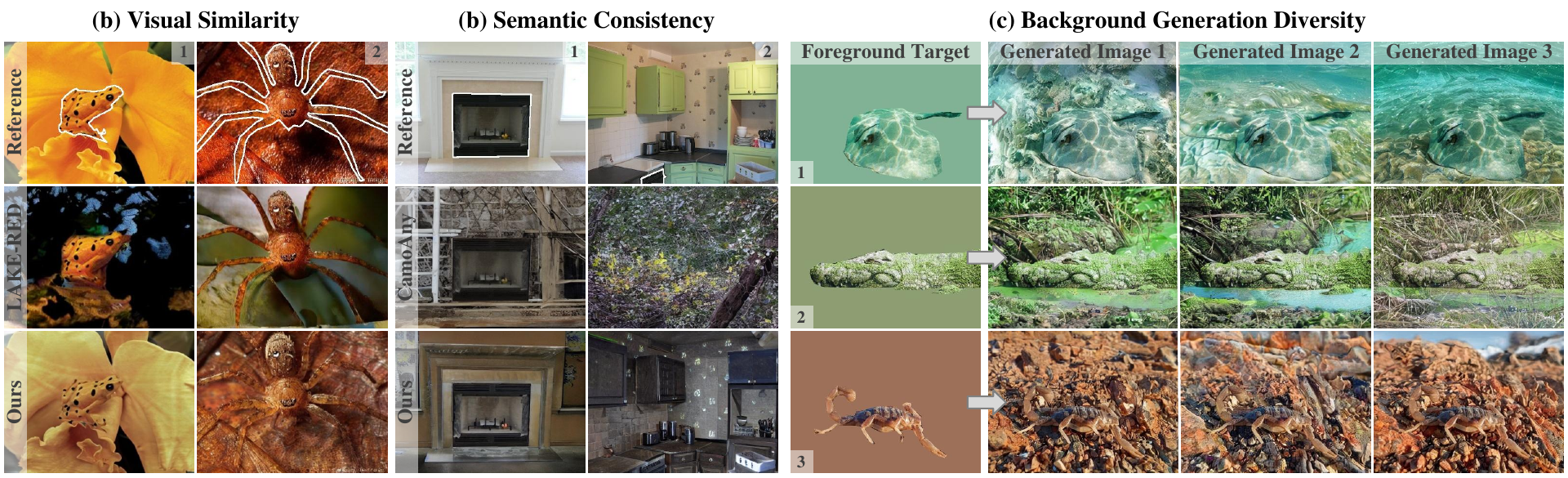}
    \captionof{figure}{\textbf{Motivation and qualitative comparison of camouflaged image synthesis.}
    \textbf{(a) LAKE-RED \cite{CVPR2024LAKERED}} fails to achieve effective camouflage, with targets remaining visually distinguishable, whereas our method produces images with well-camouflaged targets.
    \textbf{(b) CamoAny \cite{CVPR2025CamoAny}} achieves visual camouflage but suffers from weak semantic consistency between the target and the generated background; in contrast, our method preserves semantic coherence, resulting in more realistic scenes.
    \textbf{(c) Our approach} is capable of generating diverse and realistic camouflaged images while jointly satisfying visual similarity and semantic consistency.}
    \label{Fig:Motivation}
  \end{center}
}
]



\printAffiliationsAndNotice{}  

\begin{abstract}
Camouflaged image generation (CIG) has recently emerged as an efficient alternative for acquiring high-quality training data for camouflaged object detection (COD). However, existing CIG methods still suffer from a substantial gap to real camouflaged imagery: generated images either lack sufficient camouflage due to weak visual similarity, or exhibit cluttered backgrounds that are semantically inconsistent with foreground targets.
To address these limitations, we propose RealCamo, a novel out-painting-based framework for controllable realistic camouflaged image generation. RealCamo explicitly introduces additional layout controls to regulate global image structure, thereby improving semantic coherence between foreground objects and generated backgrounds. Moreover, we construct a multimodal textual-visual condition by combining a unified fine-grained textual task description with texture-oriented background retrieval, which jointly guides the generation process to enhance visual fidelity and realism. To quantitatively assess camouflage quality, we further introduce a background-foreground distribution divergence metric that measures the effectiveness of camouflage in generated images. Extensive experiments and visualizations demonstrate the effectiveness of our proposed framework.

Code and results will be released.
\end{abstract}

\section{Introduction}

\textbf{Why Synthetic Data:}
Concealed visual perception tasks, exemplified by camouflaged object detection (COD) \cite{ICML2025RUN}, aim to identify semantically meaningful targets that are visually indistinguishable from their surrounding backgrounds. Such tasks are increasingly important in real-world applications, including medical diagnosis, agricultural and industrial production \cite{AIR2024CODsurvey}. Despite recent progress, the development of COD remains fundamentally constrained by the limited scale of available training data.

In contrast to large-scale benchmarks for general object detection \& segmentation, such as PASCAL VOC \cite{IJCV2010PASCAL} and MS-COCO \cite{ECCV2014MScoco}, which contain tens to hundreds of thousands of precisely annotated images, existing COD datasets \cite{CVPR2020SINet-COD10K, CVPR2021LSR-NC4K} typically comprise only a few thousand samples. This scarcity arises from two intrinsic challenges: \textbf{(i)} camouflaged objects are inherently difficult to annotate due to their high visual similarity to backgrounds, and \textbf{(ii)} camouflaged scenes with high quality are themselves rare and costly to capture in the real world. As a result, data insufficiency has become a critical bottleneck for improving both the robustness and generalization of COD models.

Recent advances in image generation models, including GAN \cite{ACM2020GAN} and diffusion models \cite{NIPS2020DDPM, CVPR2022LDM}, offer a promising alternative, while leveraging high-quality synthetic data to augment training has been shown to be an effective and cost-efficient strategy in multiple vision tasks such as salient object detection \cite{arXiv2025S3OD}. This observation motivates the camouflaged image synthesis task, whose core objective is to generate realistic and diverse camouflaged scenes that can meaningfully complement limited real-world datasets and thereby improve downstream COD performance.

\begin{figure}[t]
\centerline{\includegraphics[width=1.0\columnwidth]{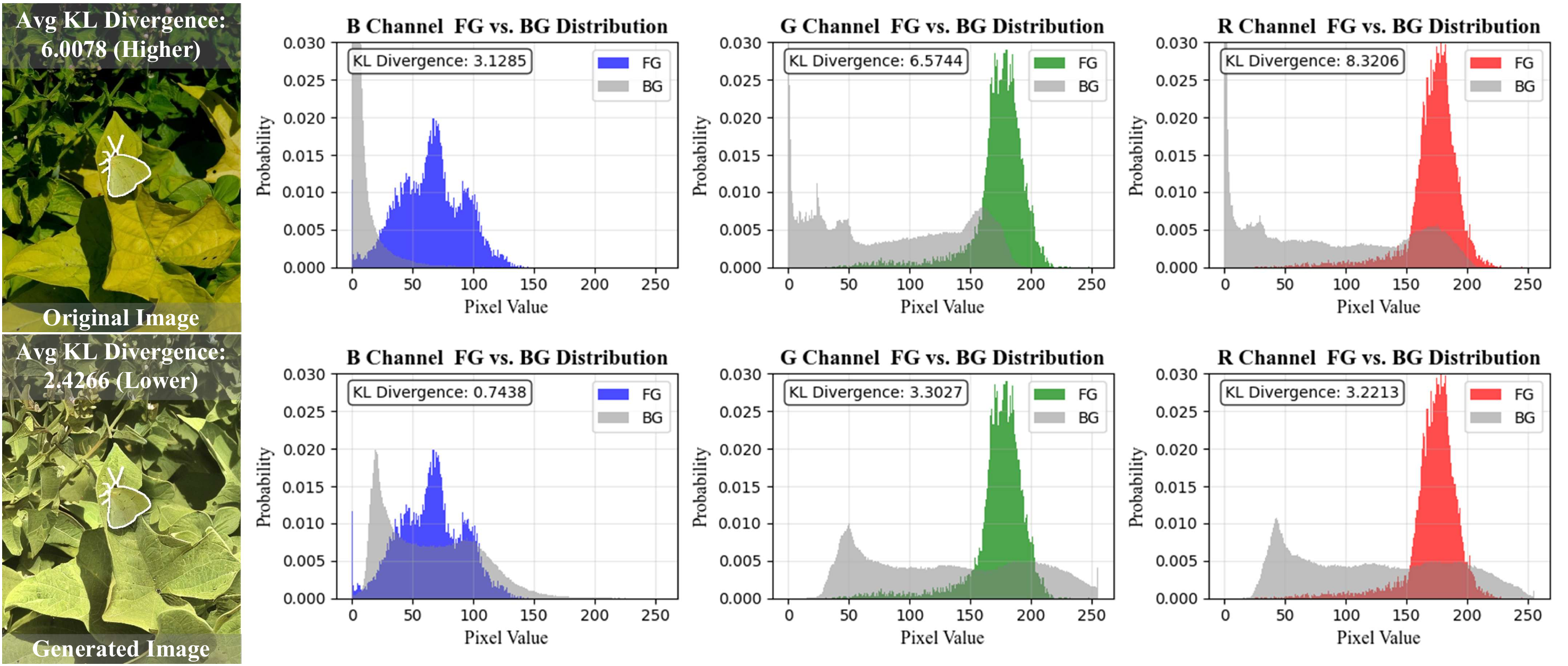}}
\caption{\textbf{The proposed KL divergence-based metric.}
Existing generative evaluation metrics, such as FID and KID, are insufficient to measure the similarity between foreground and background, while our proposed $\mathrm{KL}_{BF}$ provides a quantitative assessment of camouflage.
\textbf{Top:} Image with a higher $\mathrm{KL}_{BF}$.
\textbf{Bottom:} Our generated image with a lower $\mathrm{KL}_{BF}$, more camouflaged.}
\label{fig:KL}
\end{figure}

\textbf{Why Out-Painting:}
Existing approaches to camouflaged image synthesis can be broadly categorized into two paradigms. The first paradigm, exemplified by style transfer-based methods such as LCGNet \cite{TMM2022LCGNet}, modifies the appearance of a target object to match a given background. However, this process often alters object-specific textures and visual cues to such an extent that the semantic identity of the target is partially or entirely compromised. Consequently, these methods achieve camouflage by enforcing visual uniformity, rather than by modeling the natural interaction between an object and its environment.

The second paradigm, represented by out-painting-based methods such as LAKE-RED \cite{CVPR2024LAKERED}, follows a fundamentally different strategy. Instead of altering the target object, these methods preserve its semantic content and generate surrounding background regions whose textures and structures are consistent with the target appearance. This formulation enables camouflage to emerge from contextual compatibility between the object and its environment, rather than from direct manipulation of the object itself.

Compared with style transfer-based approaches, out-painting offers several advantages. First, it maintains the semantic integrity of the target while avoiding the need for an externally provided background image. More importantly, it aligns more closely with the natural principles of camouflage observed in the real world. Camouflage is inherently a relative and environment-dependent phenomenon: the same object may appear well camouflaged in one context but highly salient in another \cite{arXiv2025COD-SOD}. From the perspective of \textit{natural selection}, creatures do not actively redesign their appearance to match each environment; instead, environmental variation favors the survival of individuals whose appearances are already compatible with their surroundings. This perspective naturally corresponds to the out-painting formulation.

\textbf{Why Background Control:}
For out-painting-based methods to produce realistic and effective camouflaged training data, we argue that two key requirements must be simultaneously satisfied. First, the generated background should exhibit strong appearance compatibility with the target, particularly in terms of color distribution and texture patterns, so that the target is genuinely camouflaged under visual perception. Second, the background must convey semantically plausible content consistent with real-world priors; for instance, aquatic objects should appear in water rather than in visually incompatible contexts.

Meeting these requirements is inherently challenging because camouflaged targets typically occupy only a small spatial region of the image. Consequently, out-painting methods must synthesize large background areas with limited target-conditioned information, often leading to a trade-off between visual similarity and semantic consistency. As shown in Fig. \ref{Fig:Motivation} (a \& b), existing approaches exhibit clear deficiencies: some generated backgrounds that are insufficiently compatible with the target appearance, resulting in weak camouflage, while others produce visually plausible textures that lack coherent or meaningful semantics.

These limitations indicate that unconstrained out-painting is inadequate for high-fidelity camouflaged image generation. To address this issue, explicit guidance and control mechanisms during background synthesis are essential, enabling the generation process to balance appearance-level camouflage with semantic realism.

\textbf{Our Focus and Contributions:}
Based on the above analysis, this work focuses on three tightly coupled contributions:

\textbf{(1) Controlled camouflaged image synthesis.}
To overcome the limitations of existing out-painting-based methods in background generation, we introduce a novel out-painting-based framework termed \textbf{RealCamo}, which integrates \textbf{explicit layout controls} and \textbf{textual-visual guidance}. We argue that background layout plays a critical role in determining semantic plausibility. Building upon the contrast control in CamoAny, we incorporate additional depth and holistically-nested edge information into ControlNet to explicitly constrain the spatial layout of generated scenes, thereby preserving semantic realism. Furthermore, to enhance camouflage effectiveness at the texture level, we propose a textual–visual guidance mechanism that combines a unified fine-grained textual task description with additional visual cues retrieved according to target texture characteristics, enhancing visual fidelity and realism of generation.

\textbf{(2) Camouflage effectiveness evaluation metrics.}
A realistic camouflaged image should preserve the original scene layout while achieving camouflage through modifications in background color and texture. Existing generative evaluation metrics, such as FID and KID, are insufficient to capture this requirement. We therefore introduce SSIM to quantify structural consistency between generated and original images as an indicator for semantic realism. In addition, we propose a KL divergence-based metric \textbf{$\mathrm{KL}_{BF}$}, which measures the similarity between background and foreground pixel distributions, providing a principled quantitative assessment of camouflage effectiveness, as shown in Fig. \ref{fig:KL}.

\textbf{(3) Downstream evaluation.}
To assess the practical utility of the generated camouflaged images, we construct a large-scale synthetic dataset, \textbf{SynCOD12K}, as supplementary training data for camouflaged object detection (COD). Extensive experiments on four COD benchmarks demonstrate consistent performance improvements, with further investigations on the effects of camouflage-augmented data on related salient and general object detection tasks.

\section{Related Work}

\textbf{Synthetic Dataset Generation.}
Synthetic data has become an effective strategy for alleviating data scarcity in computer vision \cite{MIR2024Synsurvey}. Advances in generative models,
including GANs and diffusion models,
have enabled the construction of large-scale, high-quality synthetic datasets.
DatasetGAN \cite{CVPR2021DatasetGAN} generates image-annotation pairs by exploiting the latent feature space of pretrained GANs,
while BigDatasetGAN \cite{CVPR2022BigDatasetGAN} extends this paradigm to ImageNet-scale class diversity using VQGAN \cite{CVPR2021VQGAN}.
More recently, diffusion-based methods such as DiffuMask \cite{ICCV2023DiffuMask} and DatasetDM \cite{CVPR2022LDM} leverage text-to-image models to produce photorealistic images and corresponding annotations for downstream perception tasks.
Despite their success, these approaches primarily focus on generic image synthesis and annotation, and remain limited when applied to structurally complex and semantically subtle scenarios such as camouflaged image generation.

\textbf{Camouflaged Image Generation.}
Camouflaged image generation, introduced by \citet{TOG2010Chu}, can be categorized into two paradigms:
\textbf{\textit{Style transfer-based background-guided}} approaches modify foreground appearance to blend targets into given backgrounds.
DCI \cite{AAAI2020DCI} employs an attention-aware camouflage loss to suppress salient cues while preserving subtle perceptual features, and LCGNet \cite{TMM2022LCGNet} achieves extreme camouflage by fusing high-level foreground and background representations.
In contrast, \textbf{\textit{out-painting-based foreground-guided methods}} generate backgrounds conditioned on the foreground target.
LAKE-RED \cite{CVPR2024LAKERED} introduces a knowledge retrieval–augmented framework using a pretrained codebook, while FACIG \cite{ICME2025FACIG} reduces foreground distortion with a foreground-aware denoising loss.
CamoAny \cite{CVPR2025CamoAny} further improves background synthesis through controlled out-painting and representation engineering. Despite these advances, existing methods still struggle to simultaneously achieve strong visual similarity and semantic consistency.
Recent text-driven method CT-CIG \cite{AAAI2026CTCIG} has shown promise in realistic camouflage synthesis, but the inability to provide aligned annotations limits its applicability to downstream camouflaged object detection applications.

\textbf{Controlled Image Generation.}
Recent advances in text-to-image (T2I) generation integrate large-scale text encoders, such as CLIP \cite{ICML2021CLIP} and T5 \cite{JMLR2020T5}, into diffusion models, yielding powerful T2I models. To enable fine-grained spatial control, ControlNet \cite{ICCV2023ControlNet} introduces an auxiliary trainable branch to inject conditional signals into pretrained diffusion models. Building upon this design, UniControl \cite{NIPS2023UniControl} employs a task-aware hypernetwork to modulate zero convolutions of the ControlNet in handling multiple conditions, while CtrLoRA \cite{ICLR2025CtrLoRA} further reduces parameter overhead through condition-specific LoRA \cite{ICLR2022LoRA} fine-tuning, and \citet{ICCV2025PreserveAnything} proposes an N-channel ControlNet to improve object preservation, prompt alignment, and aesthetic quality under multiple control settings.

\section{Methodology}

\subsection{Preliminary}
Our proposed method is based on the classic Latent Diffusion Models (LDMs) \cite{CVPR2022LDM} and ControlNet \cite{ICCV2023ControlNet}. Here, we briefly revisit their basics, with more details can be found in Appendix \ref{Appendix:Preliminary}.

\begin{figure*}[t]
\centerline{\includegraphics[width=1.0\textwidth]{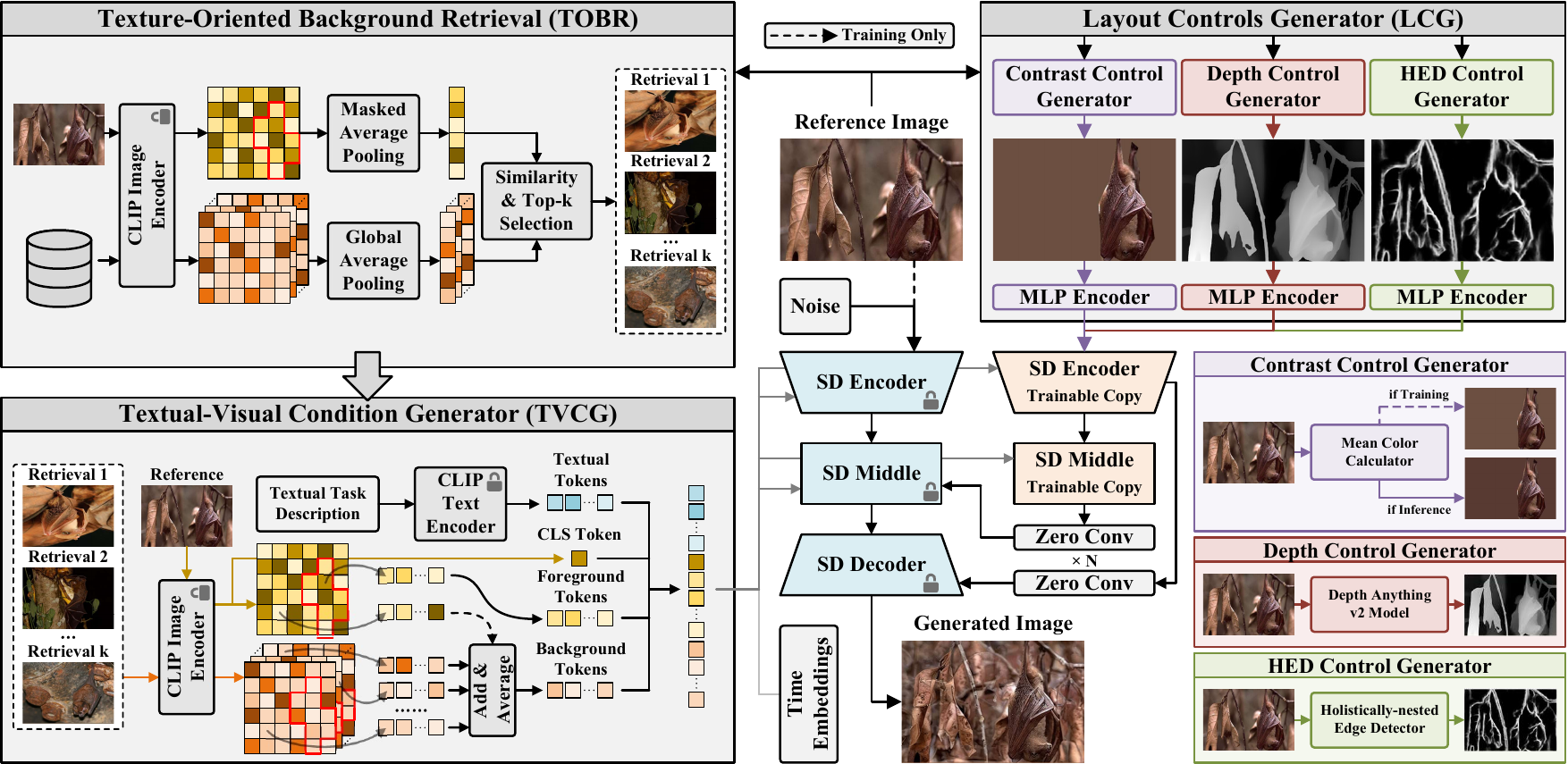}}
\caption{\textbf{The pipeline of our camouflaged image generation framework RealCamo.} Explicit layout controls (\textit{i.e.}, the contrast, depth, and HED controls) are extracted by the \textbf{Layout Controls Generator} and injected into ControlNet, while the multimodal condition produced by the \textbf{Textual-Visual Condition Generator} is used as guidance in refining the camouflage generation process.}
\label{fig:Main}
\end{figure*}

\textbf{Latent Diffusion Models.} LDMs generate high-quality images by reversing a Markov forward process in the latent space. Specifically, Given a VAE \cite{ICML2014VAE2} with encoder $\mathcal{E}$ and decoder $\mathcal{D}$, Gaussian noise $\epsilon$ is progressively added to a latent representation $z_0 = \mathcal{E}(x_0)$ of the original image $x_0$, resulting in a noisy image $z_t = \alpha_t z_0 + (1 - \alpha_t) \epsilon$, with more noise being introduced when $\alpha_t$ decreases as timestep $t$ increases. A denoising network $\epsilon_{\theta}$ is trained to predict the added noise given a condition $c^t$ following:
\begin{equation}
\label{Equation:DDPM}
    \mathcal{L} = \mathbb{E}_{x_0, t, \epsilon \sim \mathcal{N}(0, I), \mathcal{E}} \left[ \parallel \epsilon - \epsilon_{\theta}(z_t, t, c^t) \parallel \right],
\end{equation}
and reverse the noise image to a latent representation $z'_0$, which is finally decoded into the image space by $\mathcal{D}(z'_0)$.

\textbf{ControlNet.}
ControlNet extends the structural controllability of LDMs by incorporating additional conditions ($c^{f}$) like Canny edges and human pose. With the introduction of an additional trained branch aided by zero convolutions, ControlNet reformulates Equation \ref{Equation:DDPM} as:
\begin{equation}
\label{Equation:ControlNet}
    \mathcal{L} = \mathbb{E}_{x_0, t, c^t, c^f, \epsilon \sim \mathcal{N}(0. I), \mathcal{E}} \left[ \parallel \epsilon - \epsilon_{\theta}(z_t, t, c^t, c^f) \parallel_2^2 \right],
\end{equation}
which enables fine-grained spatial control during generation.

\subsection{Framework Overview}
As illustrated in Fig. \ref{fig:Main}, the proposed \textbf{RealCamo} framework is built upon Stable Diffusion 1.5 \cite{CVPR2022LDM} and ControlNet \cite{ICCV2023ControlNet}, enabling controllable camouflaged image generation through complementary structural and semantic conditioning. Control signals are injected in two ways: \textbf{(i)} structural constraints introduced via ControlNet to regulate spatial layout, and \textbf{(ii)} multimodal prompts injected through cross-attention to guide image content. These mechanisms align with our objective that high-quality camouflage requires both \textbf{visual similarity} and \textbf{semantic consistency} between foreground targets and background scenes. Accordingly, we employ a Layout Controls Generator to provide explicit structural priors via ControlNet, and a Textual–Visual Condition Generator that integrates a unified fine-grained task description with background images retrieved by a Texture-Oriented Background Retrieval module, producing informative multimodal prompts to effectively steer the camouflage generation process.

\subsection{Layout Controls Generator}
We argue that semantic consistency in camouflaged image generation is largely determined by the structural layout of the image. Prior methods \cite{CVPR2024LAKERED, CVPR2025CamoAny} often yield cluttered or physically implausible backgrounds due to the lack of explicit layout constraints. To address this, we propose a Layout Controls Generator (LCG) that extracts complementary structural cues and injects them via ControlNet, as shown in Fig. \ref{fig:Main}. Specifically, for a given image $\mathrm{I}$ and a corresponding mask $\mathrm{M}$ which indicates the foreground target, we employ LCG to extract \textbf{contrast}, \textbf{depth}, and \textbf{holistically-nested edge (HED)} controls as explicit structural priors to promote coherent spatial organization and improved foreground–background semantic alignment:
\begin{equation}
    c_{\mathrm{cst}}, c_{\mathrm{dep}}, c_{\mathrm{hed}} = LCG(\mathrm{I}, \mathrm{M}, is\_training).
\end{equation}
Each control signal is embedded using a dedicated MLP \cite{NIPS2023UniControl} due to distributional differences, and then fused as structural guidance for ControlNet to process.

\textbf{Contrast Control.} Following CamoAny, contrast control $c_{\mathrm{cst}}$ aims to preserve the foreground target while providing basic regulation of color contrast, which varies when training and inference based on $is\_training$.

\textbf{Depth Control.} Depth control $c_{\mathrm{dep}}$ is generated using Depth Anything V2 \cite{NIPS2024DepthAnythingV2}, and is intended to constrain the spatial relationships within the image.

\textbf{HED Control.} HED control $c_{\mathrm{hed}}$ is obtained by deploying Holistically-nested edge detection \cite{ICCV2015HED} to regulate the contours of the main elements in the image.

\subsection{Texture-Oriented Background Retrieval}
Compared to generic image synthesis, camouflaged image generation requires backgrounds that are quite complex and sometimes unconventional. Although LAKE-RED \cite{CVPR2024LAKERED} retrieves background embeddings from a pretrained codebook, such representations learned from generic image distributions lack camouflage-specific priors. We therefore introduce a Texture-Oriented Background Retrieval (TOBR) module that uses camouflaged images as a domain-specific background knowledge base. As illustrated in Algorithm \ref{alg:TOBG}, given a target foreground object, a CLIP image encoder is adopted to extract texture-aware features for the object and all candidates in the knowledge base. Then, the top-$k$ most similar images in the embedding space are retrieved and incorporated into multimodal prompt construction to guide the camouflage generation.

\subsection{Textual-Visual Condition Generator}
Textual conditioning is central to T2I generation. Although \citet{CVPR2025CamoAny} enhances textual prompts via prompt engineering, the incorporated background semantics remain limited. While detailed scene descriptions have been shown to improve the realism of camouflaged image synthesis \cite{AAAI2026CTCIG, arXiv2026GenCAMO}, acquiring scene-specific fine-grained descriptions is often labor-intensive. We therefore propose a Textual–Visual Condition Generator (TVCG) that constructs a multimodal prompt by combining a unified fine-grained task description with high-quality background visual embeddings. This design offers richer guidance for generation and improves visual similarity between the foreground target and the background. The multimodal prompt comprises three components: \textbf{(i)} fine-grained textual task description, \textbf{(ii)} object category information, and \textbf{(iii)} reference background visual cues.

\textbf{Fine-Grained Textual Task Description.}
Instead of using overly simplistic prompts (\textit{e.g.}, \textit{``image of a fish Camouflaged with Background''} \cite{CVPR2025CamoAny}) or labor-intensive, scene-specific descriptions (\textit{e.g.}, \textit{``A fish with mottled green and brown patterns and elongated appendages in a coral reef ecosystem.''} \cite{AAAI2026CTCIG}), we adopt a unified fine-grained textual task description \textbf{\textit{``A realistic image of an object blending into its surroundings, where the background shares similar colors, textures, and patterns with the object, making it hard to distinguish. Natural lighting, photorealistic, seamless camouflage, high detail.''}} as the textual guidance in generation.
This unified description improves generation quality while eliminating the need for manually crafting scene-specific prompts, and is encoded into textual tokens $c_{\mathrm{txt}}$ via a CLIP text encoder.

\textbf{Object Category Information.}
For the given image, we adopt the \textbf{[CLS]} token processed by a CLIP image encoder as a compact category representation $c_{\mathrm{cls}}$ of the foreground object, avoiding explicit class identification while preserving informative semantic guidance.

\begin{algorithm}[t]
\caption{Texture-Oriented Background Retrieval}
\label{alg:TOBG}
\begin{algorithmic}
  \STATE {\bfseries Input:} image $\mathrm{I}$, mask $\mathrm{M}$, knowledge base with $K$ candidates $\mathrm{B}=\{ S_i \}_{i=1}^{K}$, number of retrieval samples $k$
  \STATE {\bfseries Initialize:} similarity scores $\mathrm{S} = \varnothing$, retrieval samples $\mathrm{R} = \varnothing$, target $E_t = MaskedAvgPool(CLIP_{I}(\mathrm{I}), \mathrm{M})$
  \FOR{$i=1$ {\bfseries to} $K$}
    \STATE sample $E_i = GlobalAvgPool(CLIP_{I}(S_i))$
    \STATE score $R_i = CosineSimilarity(E_t, E_i)$, $\mathrm{S} \leftarrow R_i$
  \ENDFOR
  \FOR{$j=1$ {\bfseries to} $k$}
    \STATE index $x =ArgMax(\mathrm{S})$, $\mathrm{R} \leftarrow S_x$
    \STATE {\bfseries remove} $S_x$ {\bfseries from} $\mathrm{S}$
  \ENDFOR
  \STATE {\bfseries Return:} retrieval samples $\mathrm{R} = \{ S_j \}_{j=1}^{k}$
\end{algorithmic}
\end{algorithm}

\textbf{Reference Background Visual Cues.}
To provide richer background guidance for the camouflage generation process, we incorporate the samples retrieved by TOBR as visual reference knowledge into the multimodal prompt. Specifically, we employ a CLIP image encoder to extract visual embeddings from the input image $\mathrm{I}$ and its top-$k$ TOBR-retrieved samples $\{ S_j \}_{j=1}^{k}$. Guided by the foreground mask $\mathrm{M}$, the foreground region of the input image is preserved, while the background regions are fused with the retrieved visual knowledge to construct background-aware visual cues $c_{\mathrm{vis}}$. This process can be formulated as:
\begin{equation}
    e_t = CLIP_{I}(\mathrm{I}), \ \ e_j = CLIP_{I}(S_j), \ j= 1, 2, ..., k,
\end{equation}
\begin{equation}
	c_{\mathrm{vis}}=\left\{
	\begin{aligned}
		&e_t \cdot \mathrm{M} + \frac{(\mathbf{1}-\mathrm{M})}{k+1} \cdot (e_t + \sum_{j=1}^{k}e_j), \ \textbf{if} \ \ training \\
		&e_t \cdot \mathrm{M} + \frac{(\mathbf{1}-\mathrm{M})}{k} \cdot \sum_{j=1}^{k}e_j, \ \textbf{if} \ \ inference.
	\end{aligned}
	\right.
\end{equation}
The final multimodal conditional prompt is the concatenation of $c_{\mathrm{txt}}$, $c_{\mathrm{cls}}$, and $c_{\mathrm{vis}}$ along the sequence dimension.

\begin{figure*}[t]
\centerline{\includegraphics[width=1.0\textwidth]{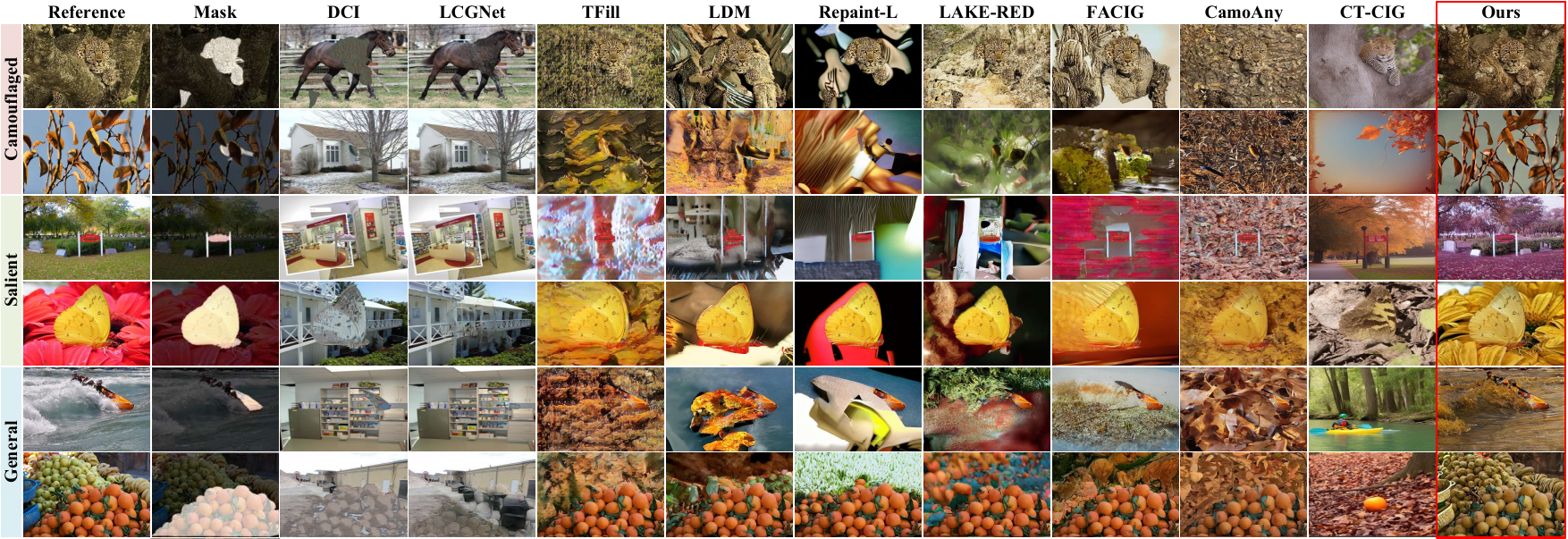}}
\caption{\textbf{Qualitative comparison with existing SOTA methods} including style transfer-based (column 3 to 4), out-painting-based (column 5 to 10), and text-driven (column 11) approaches. \textbf{Our results are demonstrated in the rightmost column with \textcolor{red}{red} box.}}
\label{fig:Result}
\end{figure*}

\begin{table*}[htbp]
\begin{center}
\caption{\textbf{Quantitative comparison with SOTA methods on LAKE-RED datasets.} The \textcolor{myred}{\textbf{1$^{st}$}}, \textcolor{mygreen}{\textbf{2$^{nd}$}} and \textcolor{myblue}{\textbf{3$^{rd}$}} best results are highlighted in \textcolor{myred}{\textbf{red}}, \textcolor{mygreen}{\textbf{green}} and \textcolor{myblue}{\textbf{blue}}. $\uparrow$ / $\downarrow$ represents the higher / lower the better. \textit{ST}: Style transfer-based. \textit{OP}: Out-paining-based. \textit{TD}: Text-driven.}
\label{tab:result}
\renewcommand{\arraystretch}{1.2}
\setlength{\tabcolsep}{3pt}
\resizebox{1.0\textwidth}{!}{
\begin{tabular}{l|c|cccc|cccc|cccc|cccc}
\hline
\multirow{2}{*}{\textbf{Method}} &
\multirow{2}{*}{\textbf{Type}} &
\multicolumn{4}{c|}{\textbf{Camouflaged Objects (6,473)}} &
\multicolumn{4}{c|}{\textbf{Salient Objects (6,473)}} &
\multicolumn{4}{c|}{\textbf{General Objects (6,473)}} &
\multicolumn{4}{c}{\textbf{Overall (19,419)}} \\
\cline{3-18}
& &
FID$\downarrow$ & KID$\downarrow$ & SSIM$\uparrow$ & $\mathrm{KL}_{BF}\downarrow$ &
FID$\downarrow$ & KID$\downarrow$ & SSIM$\uparrow$ & $\mathrm{KL}_{BF}\downarrow$ &
FID$\downarrow$ & KID$\downarrow$ & SSIM$\uparrow$ & $\mathrm{KL}_{BF}\downarrow$ &
FID$\downarrow$ & KID$\downarrow$ & SSIM$\uparrow$ & $\mathrm{KL}_{BF}\downarrow$ \\
\hline

\textbf{Original} & -- &
-- & -- & -- & 1.0027 & -- & -- & -- & 2.4837 & -- & -- & -- & 1.6818 & -- & -- & -- & 1.7227 \\
\hline

\textbf{AB} $_{\textit{IPOL 2016}}$ & \textit{ST} &
117.11 & 0.0645 & 0.2586 & 0.9200 & 126.78 & 0.0614 & 0.2753 & 1.2377 & 133.89 & 0.0645 & 0.2884 & 1.1668 & 120.21 & 0.0623 & 0.2741 & 1.1082 \\

\textbf{CI} $_{\textit{TOG 2010}}$ & \textit{ST} &
124.49 & 0.0662 & 0.2133 & 2.0678 & 136.30 & 0.7380 & 0.2117 & 2.2911 & 137.19 & 0.0713 & 0.2228 & 2.4039 & 128.51 & 0.0693 & 0.2159 & 2.2543 \\

\textbf{AdaIN} $_{\textit{Iccv 2017}}$ & \textit{ST} &
125.16 & 0.0721 & 0.1754 & 0.8821 & 133.20 & 0.0702 & 0.1880 & 1.3065 & 136.93 & 0.0714 & 0.2077 & 1.1534 & 126.94 & 0.0703 & 0.1904 & 1.1140 \\

\textbf{DCI} $_{\textit{AAAI 2010}}$ & \textit{ST} &
130.21 & 0.0689 & 0.1748 & 6.8849 & 134.92 & 0.0665 & 0.1794 & 5.8858 & 137.99 & 0.0690 & 0.2010 & 6.1119 & 130.52 & 0.0673 & 0.1850 & 6.2935 \\

\textbf{LCGNet} $_{\textit{TMM 2023}}$ & \textit{ST} &
129.80 & 0.0504 & 0.1760 & 1.6329 & 136.24 & 0.0597 & 0.1781 & 1.7395 & 132.64 & 0.0548 & 0.1995 & 1.9137 & 129.88 & 0.0550 & 0.1845 & 1.7621 \\
\hline

\textbf{TFill} $_{\textit{CVPR 2022}}$ & \textit{OP} &
63.74 & 0.0336 & \textcolor{myblue}{\textbf{0.2527}} & \textcolor{mygreen}{\textbf{0.3399}} &
96.91 & 0.0453 & 0.2886 & \textcolor{mygreen}{\textbf{0.6743}} &
122.44 & 0.0747 & 0.2989 & \textcolor{mygreen}{\textbf{0.7585}} &
80.39 & 0.0438 & 0.2801 & \textcolor{mygreen}{\textbf{0.5909}} \\

\textbf{LDM} $_{\textit{CVPR 2022}}$ & \textit{OP} &
58.65 & 0.0380 & 0.2390 & 1.0277 &
107.38 & 0.0524 & \textcolor{myblue}{\textbf{0.3287}} & 1.7082 &
129.04 & 0.0748 & \textcolor{myblue}{\textbf{0.3090}} & 1.4339 &
84.48 & 0.0488 & \textcolor{myblue}{\textbf{0.2923}} & 1.3899 \\

\textbf{RePaint-L} $_{\textit{CVPR 2022}}$ & \textit{OP} &
76.80 & 0.0459 & \textcolor{mygreen}{\textbf{0.2597}} & 3.6712 &
114.96 & 0.0497 & \textcolor{mygreen}{\textbf{0.3504}} & 4.1944 &
136.18 & 0.0686 & \textcolor{mygreen}{\textbf{0.3303}} & 3.5917 &
96.14 & 0.0498 & \textcolor{mygreen}{\textbf{0.3135}} & 3.8191 \\

\textbf{LAKE-RED} $_{\textit{CVPR 2024}}$ & \textit{OP} &
39.55 & 0.0212 & 0.2208 & 0.9191 &
89.08 & 0.0435 & 0.2881 & 1.4268 &
\textcolor{myblue}{\textbf{102.45}} & 0.0624 & 0.2736 & 1.4216 &
64.33 & 0.0357 & 0.2608 & 1.2558 \\

\textbf{FACIG} $_{\textit{ICME 2025}}$ & \textit{OP} &
\textcolor{mygreen}{\textbf{27.61}} & \textcolor{myblue}{\textbf{0.0099}} & 0.2443 & 0.8953 &
\textcolor{myblue}{\textbf{82.23}} & 0.0326 & 0.3234 & 1.5152 &
\textcolor{myred}{\textbf{96.96}} & \textcolor{myblue}{\textbf{0.0504}} & 0.2983 & 1.2428 &
\textcolor{myblue}{\textbf{52.87}} & \textcolor{myblue}{\textbf{0.0229}} & 0.2887 & 1.2177 \\

\textbf{CamoAny} $_{\textit{CVPR 2025}}$ & \textit{OP} &
34.41 & 0.0129 & 0.1591 & \textcolor{myred}{\textbf{0.2948}} &
\textcolor{myred}{\textbf{79.38}} & \textcolor{myblue}{\textbf{0.0314}} & 0.2052 & \textcolor{myred}{\textbf{0.5905}} &
\textcolor{mygreen}{\textbf{99.06}} & 0.0539 & 0.1953 & \textcolor{myred}{\textbf{0.6173}} &
55.40 & 0.0248 & 0.1865 & \textcolor{myred}{\textbf{0.5009}} \\
\hline

\textbf{CT-CIG} $_{\textit{AAAI 2026}}$ & \textit{TD} &
\textcolor{myblue}{\textbf{30.59}} & \textcolor{mygreen}{\textbf{0.0085}} & 0.1972 & 1.0199 &
\textcolor{mygreen}{\textbf{81.60}} & \textcolor{myred}{\textbf{0.0231}} & 0.1910 & 1.1938 &
104.46 & \textcolor{myred}{\textbf{0.0241}} & 0.2091 & 1.4424 &
\textcolor{mygreen}{\textbf{52.88}} & \textcolor{myred}{\textbf{0.0153}} & 0.1991 & 1.2187 \\
\hline

\rowcolor{gray!15}
\textbf{RealCamo} (Ours) & \textit{OP} &
\textcolor{myred}{\textbf{14.56}} & \textcolor{myred}{\textbf{0.0021}} & \textcolor{myred}{\textbf{0.3927}} & \textcolor{myblue}{\textbf{0.4648}} &
91.26 & \textcolor{mygreen}{\textbf{0.0276}} & \textcolor{myred}{\textbf{0.4670}} & \textcolor{myblue}{\textbf{0.9238}} &
125.65 & \textcolor{mygreen}{\textbf{0.0475}} & \textcolor{myred}{\textbf{0.4285}} & \textcolor{myblue}{\textbf{0.8365}} &
\textcolor{myred}{\textbf{52.00}} & \textcolor{mygreen}{\textbf{0.0173}} & \textcolor{myred}{\textbf{0.4294}} & \textcolor{myblue}{\textbf{0.7417}} \\

\hline

\end{tabular}}
\end{center}
\end{table*}

\subsection{Training Objective}
In RealCamo, the Stable Diffusion (SD) and the CLIP text and image encoders are all frozen, leaving ControlNet as the only trainable component. The training objective follows Equation \ref{Equation:ControlNet}, where the original control $c^{f}$ is replaced by layout controls $\{ c_{\mathrm{cst}}, c_{\mathrm{dep}}, c_{\mathrm{hed}} \}$, and the textual condition $c^{t}$ is replaced by multimodal conditions $\{ c_{\mathrm{txt}}, c_{\mathrm{cls}}, c_{\mathrm{vis}} \}$. This design enables effective controllable camouflage synthesis while preserving the generalization of the pretrained SD.

\subsection{Background-Foreground Distribution Divergence}
To more effectively quantify the degree of camouflage in synthesized images, we propose a background–foreground distribution divergence metric, denoted as $\mathrm{KL}_{BF}$, based on KL divergence. It measures visual similarity between the generated background and the foreground targets via their pixel-value distributions, where lower values indicate stronger camouflage effectiveness. Specifically, given an image $\mathrm{I}$ and its foreground mask $\mathrm{M}$, we compute the KL divergence between background and foreground distributions for each RGB channel and average them to obtain $\mathrm{KL}_{BF}$. Formally, the $\mathrm{KL}_{BF}$ can be calculated as:
\begin{align}
    \mathrm{KL}_c &= KL(\mathrm{I}_c \cdot (\mathbf{1} - \mathrm{M}), \mathrm{I}_c \cdot \mathrm{M}), \\
    \mathrm{KL}_{BF} &= Avg(\{ \mathrm{KL}_c \}), \ c \in \{ R, G, B \}.
\end{align}

\begin{figure*}[t]
\centerline{\includegraphics[width=0.9\textwidth]{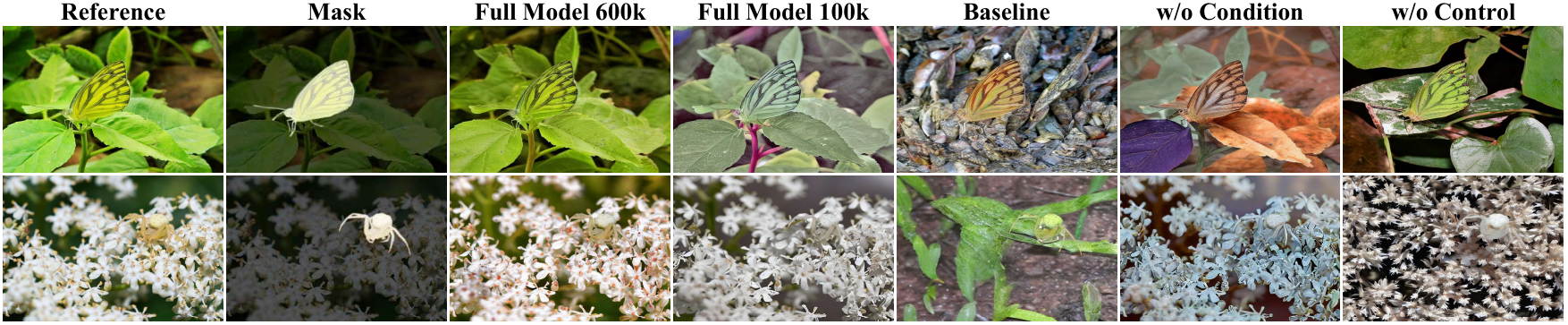}}
\caption{\textbf{Qualitative results of the ablation study on main components} of our proposed RealCamo framework.}
\label{fig:Ablation}
\end{figure*}

\begin{table*}[htbp]
\begin{center}
\caption{\textbf{Quantitative results of the ablation study on main components.} The best results are highlighted in \textbf{bold}. $\uparrow$ / $\downarrow$ represents the higher / lower the better.
For layout controls, $\times$ means only a foreground target with a white background is used, without depth and HED controls. For the textual-visual condition, $\times$ means only a simple textual prompt is used, without additional visual tokens.}
\label{tab:AblationMain}
\renewcommand{\arraystretch}{1.2}
\setlength{\tabcolsep}{3pt}
\resizebox{0.9\textwidth}{!}{
\begin{tabular}{cc|cccc|cccc|cccc|cccc}
\hline
\multicolumn{2}{c|}{\textbf{Configuration}} &
\multicolumn{4}{c|}{\textbf{Camouflaged Objects (2,000)}} &
\multicolumn{4}{c|}{\textbf{Salient Objects (2,000)}} &
\multicolumn{4}{c|}{\textbf{General Objects (2,000)}} &
\multicolumn{4}{c}{\textbf{Overall (6,000)}} \\
\hline
Layout Controls & Textual-Visual Condition &
FID$\downarrow$ & KID$\downarrow$ & SSIM$\uparrow$ & $\mathrm{KL}_{BF}\downarrow$ &
FID$\downarrow$ & KID$\downarrow$ & SSIM$\uparrow$ & $\mathrm{KL}_{BF}\downarrow$ &
FID$\downarrow$ & KID$\downarrow$ & SSIM$\uparrow$ & $\mathrm{KL}_{BF}\downarrow$ &
FID$\downarrow$ & KID$\downarrow$ & SSIM$\uparrow$ & $\mathrm{KL}_{BF}\downarrow$ \\
\hline

$\times$ & $\times$ &
39.69 & 0.0144 & 0.1446 & 0.7371 &
103.16 & 0.0372 & 0.2002 & 1.5530 &
141.34 & 0.0798 & 0.1953 & 1.3155 &
70.81 & 0.0317 & 0.1800 & 1.2019 \\

$\surd$ & $\times$ &
36.75 & 0.0130 & 0.2910 & 0.4019 &
101.28 & 0.0364 & 0.3525 & 0.9349 &
139.78 & 0.0738 & 0.3160 & 0.8637 &
65.58 & 0.0290 & 0.3198 & 0.7335 \\

$\times$ & $\surd$ &
35.42 & 0.0108 & 0.1712 & 0.7405 &
100.34 & 0.0346 & 0.2320 & 1.5997 &
138.15 & 0.0660 & 0.2186 & 1.3464 &
63.97 & 0.0268 & 0.2073 & 1.2289 \\

\rowcolor{gray!15}
$\surd$ & $\surd$ &
\textbf{31.63} & \textbf{0.0075} & \textbf{0.3254} & \textbf{0.3398} &
\textbf{98.28} & \textbf{0.0324} & \textbf{0.4027} & \textbf{0.8659} &
\textbf{136.95} & \textbf{0.0613} & \textbf{0.3695} & \textbf{0.7938} &
\textbf{60.40} & \textbf{0.0243} & \textbf{0.3659} & \textbf{0.6665} \\
\hline

\end{tabular}}
\end{center}
\end{table*}

\begin{table}[htbp]
\begin{center}
\caption{\textbf{Quantitative results of ablations on layout controls.} The best results are highlighted in \textbf{bold}. $\uparrow$ / $\downarrow$ represents the higher / lower the better.}
\label{tab:AblationCtrl}
\renewcommand{\arraystretch}{1.2}
\setlength{\tabcolsep}{3pt}
\resizebox{0.8\columnwidth}{!}{
\begin{tabular}{ccc|cccc}
\hline
\multicolumn{3}{c|}{\textbf{Configuration}} &
\multicolumn{4}{c}{\textbf{Overall (6,000)}} \\
\hline
Contrast & Depth & HED &
FID$\downarrow$ & KID$\downarrow$ & SSIM$\uparrow$ & $\mathrm{KL}_{BF}\downarrow$ \\
\hline

$\surd$ & $\times$ & $\times$ & 63.73 & 0.0272 & 0.1917 & \textbf{0.6343} \\
$\surd$ & $\surd$ & $\times$ & 61.53 & 0.0258 & 0.2887 & 0.6994 \\
$\surd$ & $\times$ & $\surd$ & 61.25 & 0.0250 & 0.2949 & 0.6957 \\

\rowcolor{gray!15}
$\surd$ & $\surd$ & $\surd$ & \textbf{60.40} & \textbf{0.0243} & \textbf{0.3659} & 0.6665 \\
\hline

\end{tabular}}
\end{center}
\end{table}

\section{Experiments}
\subsection{Experimental Settings}
\textbf{Datasets and Metrics.}
We conduct experiments on the LAKE-RED dataset \cite{CVPR2024LAKERED}, which contains 4,040 training images and 19,419 images for evaluation, including 6,473 camouflaged, salient, and general images, respectively. Following \citet{CVPR2024LAKERED}, we adopt FID \cite{ICLR2018FID} and KID \cite{NIPSl2017KID} to assess the overall quality of generated images. In addition, SSIM \cite{TIP2004SSIM} is employed to measure structural consistency between generated images and their original references, while our proposed $\mathrm{KL}_{BF}$ metric is used to quantify the effectiveness of visual camouflage.

\textbf{Implementation Details.}
The proposed framework is built upon Stable Diffusion 1.5 \cite{CVPR2022LDM} augmented with ControlNet \cite{ICCV2023ControlNet}. During training, all input images, layout controls, and corresponding masks are resized to $512 \times 512$, and a batch size of 4 is used. We set the control scale to 1.0 and train the model for 600,000 iterations with a learning rate of 1e-5. The generated images are also in a resolution of $512 \times 512$ and are resized to their original size for evaluation. All experiments are conducted on a single NVIDIA L20 GPU.

\begin{table}[htbp]
\begin{center}
\caption{\textbf{Quantitative results of ablations on textual-visual condition.} The best results are highlighted in \textbf{bold}. $\uparrow$ / $\downarrow$ represents the higher / lower the better.
\textit{Simple}: The simple textual prompt. \textit{Detailed}: The unified fine-grained textual task description.}
\label{tab:AblationCond}
\renewcommand{\arraystretch}{1.2}
\setlength{\tabcolsep}{3pt}
\resizebox{0.75\columnwidth}{!}{
\begin{tabular}{cc|cccc}
\hline
\multicolumn{2}{c|}{\textbf{Configuration}} &
\multicolumn{4}{c}{\textbf{Overall (6,000)}} \\
\hline
Textual & Visual &
FID$\downarrow$ & KID$\downarrow$ & SSIM$\uparrow$ & $\mathrm{KL}_{BF}\downarrow$ \\
\hline

\textit{Simple} & $\times$ & 65.58 & 0.0290 & 0.3198 & 0.7335 \\
\textit{Detailed} & $\times$ & 63.09 & 0.0277 & 0.3286 & 0.6962 \\
$\times$ & $\surd$ & 61.66 & 0.0260 & 0.3334 & 0.6865 \\

\rowcolor{gray!15}
\textit{Detailed} & $\surd$ & \textbf{60.40} & \textbf{0.0243} & \textbf{0.3659} & \textbf{0.6665} \\
\hline

\end{tabular}}
\end{center}
\end{table}

\subsection{Comparison with the State-of-the-arts (SOTAs)}
We compare our RealCamo with 12 SOTAs, including five style transfer-based methods (AB \cite{IPOL2016AB}, CI \cite{TOG2010Chu}, AdaIN \cite{ICCV2017AdaIN}, DCI \cite{AAAI2020DCI}, and LCGNet \cite{TMM2022LCGNet}), six out-painting-based methods (TFill \cite{CVPR2022TFill}, LDM \cite{CVPR2022LDM}, RePaint-L \cite{CVPR2022Repaint}, LAKE-RED \cite{CVPR2024LAKERED}, FACIG \cite{ICME2025FACIG}, and CamoAny \cite{CVPR2025CamoAny}), and one recently text-driven method (CT-CIG \cite{AAAI2026CTCIG}). We re-evaluate the benchmark results provided by LAKE-RED, FACIG, and CT-CIG. Due to a lack of available resources, we reproduce CamoAny with the same settings as ours.

\begin{table*}[htbp]
\begin{center}
\caption{\textbf{Performance of MVANet \cite{CVPR2024MVANet} on four COD benchmark datasets with different training data.} The best results are highlighted in \textbf{bold}. $\uparrow$ / $\downarrow$ represents the higher / lower the better.
Original (Ori.): The standard COD training data.
Gen $_{S\&G}$: Synthesis images by RealCamo of the salient and general subsets in the LAKE-RED validation set.
SynCOD12K: Synthesis images by RealCamo of the LAKE-RED training set.}
\label{tab:COD}
\renewcommand{\arraystretch}{1.2}
\setlength{\tabcolsep}{3pt}
\resizebox{0.9\textwidth}{!}{
\begin{tabular}{l|ccccc|ccccc|ccccc|ccccc}
\hline
\multirow{2}{*}{\textbf{Training Data}} &
\multicolumn{5}{c|}{\textbf{CHAMELEON (76)}} &
\multicolumn{5}{c|}{\textbf{CAMO (250)}} &
\multicolumn{5}{c|}{\textbf{COD10K (2,026)}} &
\multicolumn{5}{c}{\textbf{NC4K (4,121)}} \\
\cline{2-21}
&
$S_\alpha \uparrow$&$E_\phi \uparrow$&$F_\beta \uparrow$&$F_\beta^w \uparrow$&$M \downarrow$&
$S_\alpha \uparrow$&$E_\phi \uparrow$&$F_\beta \uparrow$&$F_\beta^w \uparrow$&$M \downarrow$&
$S_\alpha \uparrow$&$E_\phi \uparrow$&$F_\beta \uparrow$&$F_\beta^w \uparrow$&$M \downarrow$&
$S_\alpha \uparrow$&$E_\phi \uparrow$&$F_\beta \uparrow$&$F_\beta^w \uparrow$&$M \downarrow$\\
\hline

Original &
0.915 & 0.929 & 0.871 & 0.846 & 0.022 &
0.870 & 0.896 & 0.842 & 0.809 & 0.055 &
0.882 & 0.917 & 0.828 & 0.792 & 0.025 &
0.886 & 0.910 & 0.855 & 0.820 & 0.038 \\

Ori. + Gen $_{S\&G}$ &
0.905 & 0.927 & 0.846 & 0.796 & 0.031 &
0.826 & 0.842 & 0.760 & 0.717 & 0.079 &
0.865 & 0.896 & 0.784 & 0.719 & 0.035 &
0.879 & 0.897 & 0.821 & 0.769 & 0.048 \\

\rowcolor{gray!15}
Ori. + SynCOD12K  &
\textbf{0.924} & \textbf{0.949} & \textbf{0.889} & \textbf{0.871} & \textbf{0.019} &
\textbf{0.882} & \textbf{0.916} & \textbf{0.856} & \textbf{0.826} & \textbf{0.049} &
\textbf{0.896} & \textbf{0.934} & \textbf{0.850} & \textbf{0.818} & \textbf{0.021} &
\textbf{0.891} & \textbf{0.919} & \textbf{0.860} & \textbf{0.824} & \textbf{0.037} \\
\hline

\end{tabular}}
\end{center}
\end{table*}

\textbf{Qualitative Comparison.}
Fig. \ref{fig:Result} shows the qualitative comparison between our RealCamo and other methods on camouflaged, salient, and generic objects. The style transfer-based LCGNet embeds targets into given backgrounds, achieving strong visual camouflage but severely altering target semantics. In contrast, out-painting-based methods often produce insufficient camouflage or cluttered, semantically implausible backgrounds, degrading realism. Although recent text-driven CT-CIG generates visually convincing results, they fail to preserve original targets and thus require re-annotation, limiting their usefulness for downstream COD tasks. In comparison, our RealCamo achieves both high visual similarity and strong foreground–background semantic consistency while preserving target integrity, resulting in more realistic and practically valuable camouflaged images.

\textbf{Quantitative Comparison.}
Quantitative results are summarized in Tab. \ref{tab:result}. Compared to early style transfer-based methods, recent out-painting-based and text-driven approaches achieve obviously better image quality, indicated by lower FID and KID. Owing to the introduction of explicit layout controls and textual–visual multimodal conditioning, our RealCamo attains highly competitive generation performance. Although it does not achieve the lowest $\mathrm{KL}_{BF}$ score, our method balances visual camouflage with structural and semantic fidelity, producing more realistic images. By revisiting Fig. \ref{fig:Result}, methods with lower $\mathrm{KL}_{BF}$ values (\textit{e.g.}, CamoAny and TFill) indeed yield stronger visual camouflage, further validating our proposed $\mathrm{KL}_{BF}$ as an effective metric for quantifying camouflage effectiveness.
We also note that RealCamo attains higher FID under salient \& general object transfer scenes. This is largely due to the domain gap introduced by structural controls. However, this does not indicate degraded perceptual quality, while KID, extensive visualizations, and user study results consistently show that RealCamo produces high-fidelity camouflaged images.
These observations highlight the limitations of FID and KID for evaluating camouflaged image generation, which further motivates our introduction of SSIM and $\mathrm{KL}_{BF}$ metrics.

\textbf{User Study.}
We further conduct a user study to comprehensively evaluate the quality of synthesized camouflaged images based on human feedback. Detailed evaluation settings, results, and analyses are reported in Appendix \ref{Appendix:UserStudy}.

\subsection{Ablation Study}
We conduct ablation studies to validate the proposed components. Due to computational constraints, all variants are trained for 100,000 iterations and evaluated on 6,000 images uniformly sampled from three validation subsets. Results in Tab. \ref{tab:AblationMain} and Fig. \ref{fig:Ablation} show that introducing explicit layout controls improves structural consistency and overall image quality, supporting our claim that \textit{``the semantics of images are largely determined by their structural layout''}. Incorporating textual–visual condition further enhances visual realism and foreground–background visual similarity and semantic consistency. Additional results of ablations on individual layout control and textual–visual components are reported in Tabs \ref{tab:AblationCtrl} and \ref{tab:AblationCond}. More quantitative and qualitative results with further analyses are provided in Appendix \ref{Sec:App-Experiments}.

\subsection{Downstream Application}
\begin{figure}[t]
\centerline{\includegraphics[width=1.0\columnwidth]{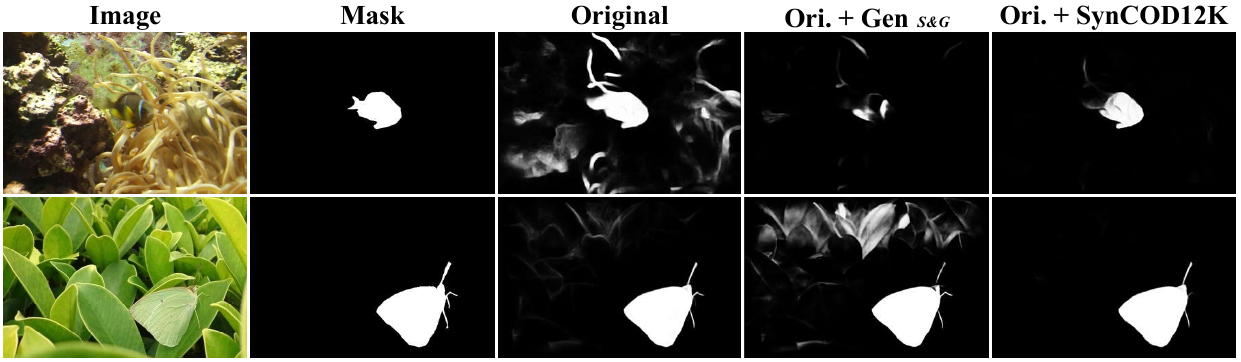}}
\caption{\textbf{Prediction results} with different training data.}
\label{fig:COD}
\end{figure}

We further evaluate the impact of synthesized camouflaged images on the downstream COD task. An awesome dichotomous segmentation model, MVANet \cite{CVPR2024MVANet}, is trained with different data configurations and evaluated on four COD benchmarks. Quantitative results and prediction results are presented in Tab. \ref{tab:COD} and Fig. \ref{fig:COD}.
We can observe that converting salient or general images into camouflaged ones degrades COD performance, possibly due to domain gaps between these synthesized samples and real COD data. In contrast, training with camouflaged-synthesis images consistently improves COD performance, highlighting the effectiveness of our generation approach for data augmentation. We also examine the impact of synthesized images on salient and general object detection. More details of experimental settings, metrics and results, and the constructed \textbf{SynCOD12K} dataset, are reported in Appendix \ref{Sec:App-DownStreamApp}.

\section{Conclusion}
In this paper, we propose RealCamo, a novel framework for controllable realistic camouflaged image generation. RealCamo explicitly incorporates additional layout controls to regulate the structural composition of generated images, thereby enhancing semantic consistency between foreground objects and synthesized backgrounds. In addition, it constructs a multimodal textual–visual condition by combining a unified fine-grained textual task description with texture-oriented background retrieved visual cues, which jointly guide the generation process to improve visual quality and realism. Extensive experiments and qualitative analyses validate the effectiveness of the proposed framework and demonstrate its strong potential for enhancing downstream camouflaged object detection applications.

\bibliography{Reference}
\bibliographystyle{icml2026}

\newpage
\appendix

\setcounter{table}{0}
\setcounter{figure}{0}
\setcounter{equation}{0}
\renewcommand{\thetable}{S\arabic{table}}
\renewcommand{\thefigure}{S\arabic{figure}}
\renewcommand{\theequation}{S\arabic{equation}}

\onecolumn

\section{Camouflaged Object Detection}
\label{Appendix:COD}
\subsection{Related Work}
As the extension of general object detection (GOD) \cite{CVPR2015FCN-GOD} and salient object detection (SOD) \cite{TIP2015SOD}, camouflaged object detection (COD) \cite{CVPR2020SINet-COD10K, ICML2025RUN} is proposed to accurately localize and precisely segment camouflaged targets hidden within complex scenes, which is applicable to numerous real-world scenarios, such as concealed defect detection in industry, pest monitoring in agriculture, and medical lesions diagnosis \cite{AIR2024CODsurvey}.
SINet \cite{CVPR2020SINet-COD10K} pioneered COD using a biologically inspired two-stage search-and-identify framework, followed by a wide range of bio-inspired \cite{CVPR2021PFNet, TPAMI2022SINetV2, CVPR2022ZoomNet, CVPR2022SegMaR, TIP2023FSNet, TIP2024IdeNet} and feature-enhanced \cite{CVPR2021MGL, CVPR2021LSR-NC4K, TIP2022FAPNet, CVPR2023FSPNet, ICLR2024ICEG, TPAMI2024CamoFormer} methods.
Subsequent studies incorporated auxiliary cues to complement RGB images, including edges and textures \cite{IJCAI2022BgNet, MIR2023DGNet, CVPR2023FEDER}, depth information \cite{ICCV2023PopNet, ACMMM2023DaCOD, ACMMM2024DSAM}, and frequency-domain representations \cite{ACMMM2023FRINet, ACMMM2023FPNet, ECCV2024FSEL}.
Representative works exploit frequency decomposition with boundary refinement \cite{CVPR2023FEDER}, depth-induced saliency enhancement \cite{ICCV2023PopNet}, and joint frequency–spatial learning \cite{ECCV2024FSEL}.
Weakly-supervised COD \cite{AAAI2023CRNet(Scribble), ECCV2024PSCOD(Point), ACMMM2025GaoShuyong}, semi-supervised COD \cite{ECCV2024CamoTeacher, MM2024SSCOD, ICASSP2025SILNet}, and unsupervised COD \cite{CVPR2025EASE, CVPR2025UCOD-DPL, ICCV2025RISE} methods are also explored.
Despite steady progress, the development of COD remains fundamentally limited by the small scale of available training datasets, which drives the development of camouflaged image generation tasks.

\subsection{Datasets and Evaluation Metrics}
\textbf{Datasets.}
Downstream application on the COD task following the standard practice of SINet \cite{CVPR2020SINet-COD10K}. Experiments are conducted on four widely used benchmark datasets: \textit{CHAMELEON} \cite{CHAMELEON}, \textit{CAMO} \cite{CVIU2019CAMO}, \textit{COD10K} \cite{CVPR2020SINet-COD10K}, and \textit{NC4K} \cite{CVPR2021LSR-NC4K}. The \textit{CHAMELEON} dataset contains 76 images, while \textit{CAMO} consists of 1,250 images across eight classes. \textit{COD10K} includes 5,066 images spanning ten super-classes, and \textit{NC4K} serves as the largest testing benchmark with 4,121 well-annotated images.
For standard COD training, we follow the common practice by using 1,000 images from \textit{CAMO} and 3,040 images from \textit{COD10K} as the training set, which also aligns with the training protocol for camouflaged image generation in \cite{CVPR2024LAKERED}. The remaining images from these two datasets, together with all images from \textit{CHAMELEON} and \textit{NC4K}, are used for evaluation.

\textbf{Evaluation Metrics.}
To thoroughly evaluate the performance, we adopt five standard metrics widely used in COD and SOD tasks: mean absolute error ($M$) \cite{MAE}, mean E-measure ($E_{\phi}$) \cite{E-measure}, mean F-measure ($F_{\beta}$) \cite{F-measure}, weighted F-measure ($F_{\beta}^{w}$) \cite{wF-measure}, and structure measure ($S_{\alpha}$) \cite{S-measure}. Better performance is indicated by lower values of $M$ and higher values of $E_{\phi}$, $F_{\beta}$, $F_{\beta}^{w}$, and $S_{\alpha}$.

\section{Preliminary}
\label{Appendix:Preliminary}
\subsection{Diffusion Models}
Tex-to-image (T2I) generation methods, represented by Stable Diffusion \cite{CVPR2022LDM}, synthesize images conditioned on textual descriptions ($c^t$) via a denoising diffusion paradigm \cite{NIPS2020DDPM}. These models can operate either in pixel space or latent space, with latent diffusion models \cite{CVPR2022LDM} offering substantially improved computational efficiency.

\textbf{Forward Process.} In diffusion models, noise sampled from the Gaussian distribution $\epsilon \sim \mathcal{N}(0, I) $ is progressively added to a clean image $x_0$, resulting in a noisy sample $x_t$ at timestep $t$:
\begin{equation}
    x_t = \sqrt{\bar{\alpha}_t} x_0 + \sqrt{1 - \bar{\alpha}_t} \epsilon,
\end{equation}
where $\alpha_t = 1 - \beta_t$, $\bar{\alpha}_t = \prod_{s=1}^t \alpha_s$, and $\{ \beta_t \}$ follows a variance schedule.

\textbf{Training Objective.} A denoising network $\epsilon_{\theta}$ is trained to predict the injected noise $\epsilon$ by minimizing the following objective:
\begin{equation}
\label{Equa:DDPM-Appendix}
    \mathcal{L} = \mathbb{E}_{x_0, t, \epsilon \sim \mathcal{N}(0, I)} \left[ \parallel \epsilon - \epsilon_{\theta}(x_t, t, c^t) \parallel_2^2 \right].
\end{equation}

\textbf{Reverse Process.}
After training the denoising network, an image $x_0$ can be gradually obtained by repeating the denoise operation $T$ times on an entire Gaussian noise $x_T \sim \mathcal{N}(0, I)$, following:
\begin{equation}
    x_{t-1} = \frac{1}{\sqrt{\bar{\alpha}_t}} \left( x_t - \frac{\beta_t}{\sqrt{1 - \bar{\alpha}_t}} \epsilon_{\theta}(x_t, t, c^t) \right) + \sigma_t \epsilon',
\end{equation}
where $t$ starts from $T$, $\epsilon' \sim \mathcal{N}(0. I)$, and $\sigma_t^2 = \beta_t$. Intermediate estimations of $x_0$ can also be directly obtained at any timestep $t$ following:
\begin{equation}
    \hat{x}_0 = \frac{x_t - \sqrt{1 - \bar{\alpha}_t} \epsilon_{\theta}(x_t, t, c^t)}{\sqrt{\bar{\alpha}_t}}.
\end{equation}

In latent diffusion models (LDMs) \cite{CVPR2022LDM}, the above forward and reverse process is conducted on latent embeddings $z = \mathcal{E}_{\mathrm{VAE}}(x)$, as a latent representation of raw RGB pixels $x$ encoded by a VAE \cite{ICLR2014VAE1, ICML2014VAE2}. And the denoised latent representation $z_0$ is further decoded into a normal RGB image using the VAE decoder, as $x_0 = \mathcal{D}_{\mathrm{VAE}}(z_0)$.

\subsection{ControlNet}
ControlNet \cite{ICCV2023ControlNet} remarkably extends the controllability of large pretrained T2I LDMs by incorporating additional conditions ($c^{f}$) like Canny edges, human pose, or depth map, which provide structural priors for the target generated image.

\textbf{Architecture.}
ControlNet controls the generating process by injecting additional conditions into the blocks of a pretrained neural network. With the definition $\mathcal{F}(\cdot; \Theta)$ of a trained neural block with parameters $\Theta$, that transform an input feature map $x$ into an output feature map $y$ as $y = \mathcal{F}(x; \Theta)$, a trainable copy of $\mathcal{F}$ is created as $\mathcal{F}_c(\cdot; \Theta_\mathrm{c})$.

This trainable copy takes an external conditioning vector $c^f$ as input, and is connected to the locked model $\mathcal{F}$ with zero convolution layers $\mathcal{Z}(\cdot; \cdot)$, which are $1 \times 1$ convolution layers with both weight and bias initialized to zeros. The ControlNet then generated the output $y_\mathrm{c}$ by two instances of zero convolutions with parameters $\Theta_{\mathrm{Z}1}$ and $\Theta_{\mathrm{Z}2}$ as:
\begin{equation}
    y_\mathrm{c} = \mathcal{F}(x; \Theta) + \mathcal{Z}(\mathcal{F}(x + \mathcal{Z}(c^f; \Theta_{\mathrm{Z}1}); \Theta_\mathrm{c}); \Theta_{\mathrm{Z}2})
\end{equation}

\textbf{Training Objective.} With the introduction of additional conditioning vector $c^f$, the overall training objective (Equation \ref{Equa:DDPM-Appendix}) is reformulated as:
\begin{equation}
    \mathcal{L} = \mathbb{E}_{x_0, t, c^t, c^f, \epsilon \sim \mathcal{N}(0. I)} \left[ \parallel \epsilon - \epsilon_{\theta}(x_t, t, c^t, c^f) \parallel_2^2 \right].
\end{equation}

This enables fine-grained spatial control during generation, allowing precise alignment between generated content and structural constraints.

\section{Experiments}
\label{Sec:App-Experiments}

\subsection{Ablations on Layout Controls}
We conduct ablation studies on individual components of the proposed layout controls, with results summarized in Tab. \ref{tab:AppendixAblationCtrl} and Fig. \ref{fig:App-Ablation}. It can be found that incorporating depth and HED controls effectively enforces structural constraints on the synthesized backgrounds, thereby enhancing semantic consistency between the foreground target and the generated scene. This structural alignment further improves the overall generation quality and visual realism. While both depth and HED controls contribute to structure-aware generation, they exhibit complementary effects: depth control produces images with clearer spatial hierarchy and depth layering, whereas HED control sharpens object boundaries and yields more distinct foreground contours. These observations highlight the necessity of jointly modeling geometric layout and edge-level structure in controllable camouflaged image generation.

\subsection{Ablations on Textual-Visual Condition}
\begin{figure*}[t]
\centerline{\includegraphics[width=1.0\textwidth]{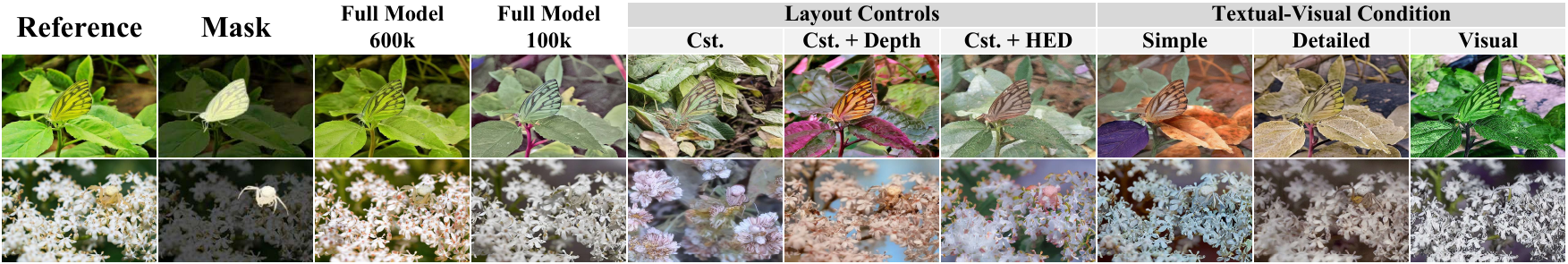}}
\caption{\textbf{Qualitative results of the ablation study on layout controls and textual-visual condition.}
Cst.: Contrast control.
Simple: The simple textual prompt.
Detailed: The unified fine-grained textual task description.}
\label{fig:App-Ablation}
\end{figure*}

\begin{figure*}[t]
\centerline{\includegraphics[width=1.0\textwidth]{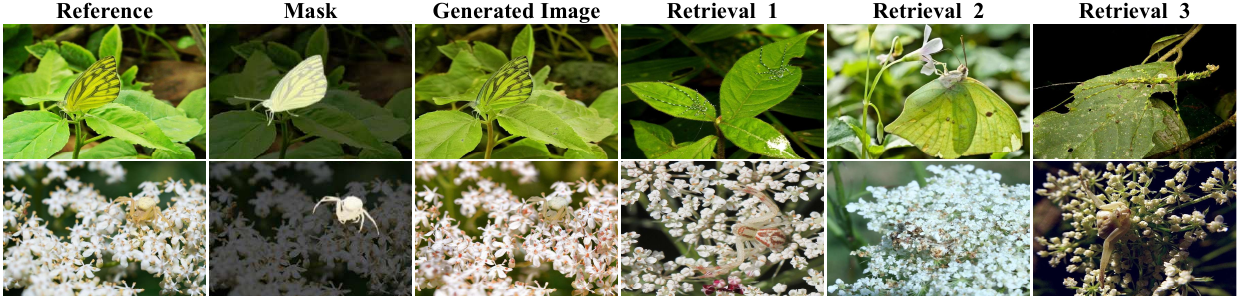}}
\caption{\textbf{Illustrations of image samples} retrieved by the proposed Texture-Oriented Background Retrieval (TOBR) module.}
\label{fig:App-Retrieval}
\end{figure*}

We conduct ablation studies on the individual components of the proposed textual–visual condition, with results reported in Tab. \ref{tab:AppendixAblationCond} and Fig. \ref{fig:App-Ablation}. \textit{Simple} and \textit{Detailed} denote the use of a simple textual prompt and the proposed unified fine-grained textual task description, respectively, while \textit{Visual} refers to the incorporation of object category information and reference background visual cues. The results show that fine-grained task descriptions consistently improve overall image generation quality. Moreover, introducing visual conditions further enhances generation fidelity by providing explicit visual guidance for background synthesis (as shown in Fig. \ref{fig:App-Retrieval}), which leads to higher visual similarity and improved semantic alignment between the foreground object and the generated background.

\begin{table*}[htbp]
\begin{center}
\caption{\textbf{Ablation results on layout controls.} The best results are highlighted in \textbf{bold}. $\uparrow$ / $\downarrow$ represents the higher / lower the better.}
\label{tab:AppendixAblationCtrl}
\renewcommand{\arraystretch}{1.2}
\setlength{\tabcolsep}{3pt}
\resizebox{1.0\textwidth}{!}{
\begin{tabular}{ccc|cccc|cccc|cccc|cccc}
\hline
\multicolumn{3}{c|}{\textbf{Configuration}} &
\multicolumn{4}{c|}{\textbf{Camouflaged Objects (2,000)}} &
\multicolumn{4}{c|}{\textbf{Salient Objects (2,000)}} &
\multicolumn{4}{c|}{\textbf{General Objects (2,000)}} &
\multicolumn{4}{c}{\textbf{Overall (6,000)}} \\
\hline
Contrast & Depth & HED &
FID$\downarrow$ & KID$\downarrow$ & SSIM$\uparrow$ & $\mathrm{KL}_{BF}\downarrow$ &
FID$\downarrow$ & KID$\downarrow$ & SSIM$\uparrow$ & $\mathrm{KL}_{BF}\downarrow$ &
FID$\downarrow$ & KID$\downarrow$ & SSIM$\uparrow$ & $\mathrm{KL}_{BF}\downarrow$ &
FID$\downarrow$ & KID$\downarrow$ & SSIM$\uparrow$ & $\mathrm{KL}_{BF}\downarrow$ \\
\hline

$\surd$ & $\times$ & $\times$ &
35.39 & 0.0103 & 0.1629 & \textbf{0.3191} &
101.54 & 0.0355 & 0.2116 & \textbf{0.8249} &
137.89 & 0.0685 & 0.2007 & \textbf{0.7589} &
63.73 & 0.0272 & 0.1917 & \textbf{0.6343} \\

$\surd$ & $\surd$ & $\times$ &
33.14 & 0.0085 & 0.2627 & 0.3565 &
100.36 & 0.0339 & 0.3299 & 0.9049 &
137.16 & 0.0665 & 0.2735 & 0.8367 &
61.53 & 0.0258 & 0.2887 & 0.6994 \\

$\surd$ & $\times$ & $\surd$ &
32.84 & 0.0082 & 0.2421 & 0.3459 &
99.01 & 0.0336 & 0.3421 & 0.8957 &
\textbf{136.54} & 0.0620 & 0.3006 & 0.8454 &
61.25 & 0.0250 & 0.2949 & 0.6957 \\

\rowcolor{gray!15}
$\surd$ & $\surd$ & $\surd$ &
\textbf{31.63} & \textbf{0.0075} & \textbf{0.3254} & 0.3398 &
\textbf{98.28} & \textbf{0.0324} & \textbf{0.4027} & 0.8659 &
136.95 & \textbf{0.0613} & \textbf{0.3695} & 0.7938 &
\textbf{60.40} & \textbf{0.0243} & \textbf{0.3659} & 0.6665 \\
\hline

\end{tabular}}
\end{center}
\end{table*}

\begin{table*}[htbp]
\begin{center}
\caption{\textbf{Quantitative results of ablation study on textual-visual condition.} The best results are highlighted in \textbf{bold}. $\uparrow$ / $\downarrow$ represents the higher / lower the better.
\textit{Simple}: The simple textual prompt. \textit{Detailed}: The unified fine-grained textual task description.}
\label{tab:AppendixAblationCond}
\renewcommand{\arraystretch}{1.2}
\setlength{\tabcolsep}{3pt}
\resizebox{1.0\textwidth}{!}{
\begin{tabular}{cc|cccc|cccc|cccc|cccc}
\hline
\multicolumn{2}{c|}{\textbf{Configuration}} &
\multicolumn{4}{c|}{\textbf{Camouflaged Objects (2,000)}} &
\multicolumn{4}{c|}{\textbf{Salient Objects (2,000)}} &
\multicolumn{4}{c|}{\textbf{General Objects (2,000)}} &
\multicolumn{4}{c}{\textbf{Overall (6,000)}} \\
\hline
Textual & Visual &
FID$\downarrow$ & KID$\downarrow$ & SSIM$\uparrow$ & $\mathrm{KL}_{BF}\downarrow$ &
FID$\downarrow$ & KID$\downarrow$ & SSIM$\uparrow$ & $\mathrm{KL}_{BF}\downarrow$ &
FID$\downarrow$ & KID$\downarrow$ & SSIM$\uparrow$ & $\mathrm{KL}_{BF}\downarrow$ &
FID$\downarrow$ & KID$\downarrow$ & SSIM$\uparrow$ & $\mathrm{KL}_{BF}\downarrow$ \\
\hline

\textit{Simple} & $\times$ &
36.75 & 0.0130 & 0.2910 & 0.4019 &
100.28 & 0.0364 & 0.3525 & 0.9349 &
139.78 & 0.0738 & 0.3160 & 0.8637 &
65.58 & 0.0290 & 0.3198 & 0.7335 \\

\textit{Detail} & $\times$ &
34.63 & 0.0111 & 0.2951 & 0.3639 &
100.78 & 0.0353 & 0.3610 & 0.9039 &
138.84 & 0.0701 & 0.3297 & 0.8209 &
63.09 & 0.0277 & 0.3286 & 0.6962 \\

$\times$ & $\surd$ &
31.68 & 0.0081 & 0.3077 & 0.3591 &
99.62 & 0.0333 & 0.3443 & 0.8862 &
\textbf{136.82} & 0.0621 & 0.3483 & 0.8142 &
61.66 & 0.0260 & 0.3334 & 0.6865 \\

\rowcolor{gray!15}
\textit{Detail} & $\surd$ &
\textbf{31.63} & \textbf{0.0075} & \textbf{0.3254} & \textbf{0.3398} &
\textbf{98.28} & \textbf{0.0324} & \textbf{0.4027} & \textbf{0.8659} &
136.95 & \textbf{0.0613} & \textbf{0.3695} & \textbf{0.7938} &
\textbf{60.40} & \textbf{0.0243} & \textbf{0.3659} & \textbf{0.6665} \\
\hline

\end{tabular}}
\end{center}
\end{table*}

\subsection{Ablations on Hyperparameters}
\begin{figure*}[htbp]
\centerline{\includegraphics[width=1.0\textwidth]{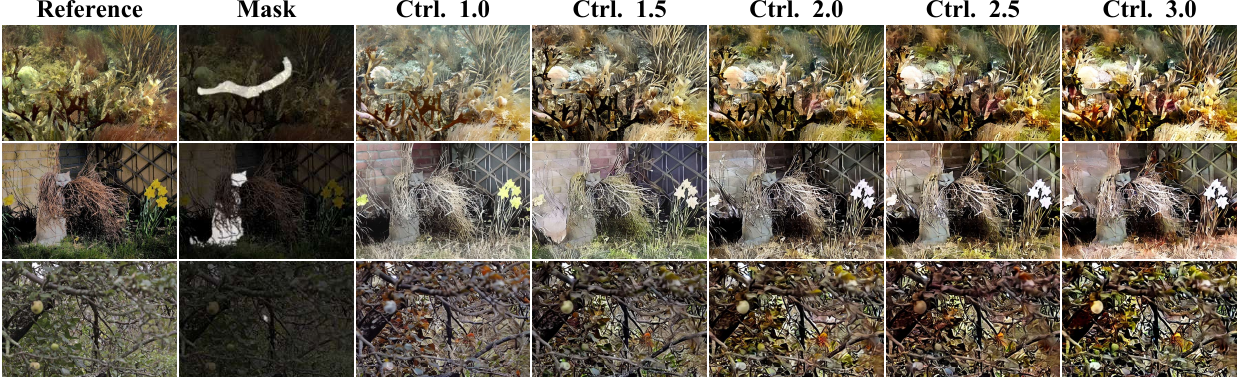}}
\caption{\textbf{Qualitative comparison of the ablation study on different control strengths.}
Ctrl.: Control strength.}
\label{fig:App-CtrlStrength}
\end{figure*}

\begin{figure*}[htbp]
\centerline{\includegraphics[width=1.0\textwidth]{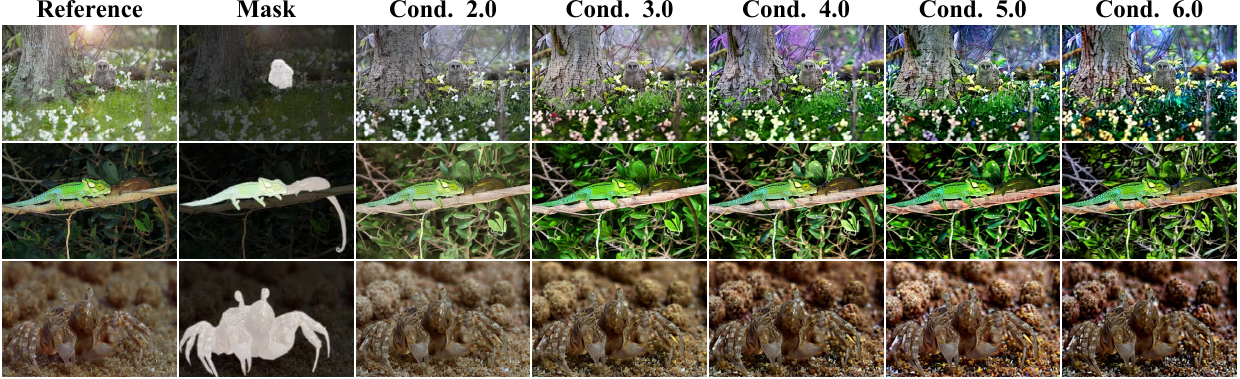}}
\caption{\textbf{Qualitative comparison of the ablation study on different condition strengths.}
Cond.: Condition strength.}
\label{fig:App-CondStrength}
\end{figure*}

We further investigate the impact of key hyperparameters on the inference behavior of the proposed framework. Specifically, we analyze the effects of \textbf{control strength} and \textbf{condition strength} as illustrated in Figs. \ref{fig:App-CtrlStrength} and \ref{fig:App-CondStrength}. As these strengths increase, the generated images exhibit more saturated colors and stronger contrast, with the overall visual appearance gradually shifting from photorealism toward a painterly style.

\begin{figure*}[htbp]
\centerline{\includegraphics[width=1.0\textwidth]{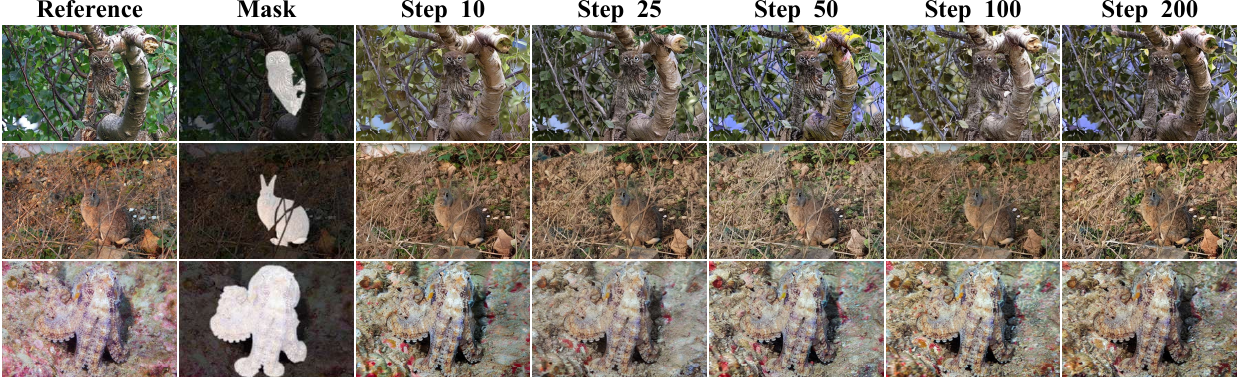}}
\caption{\textbf{Qualitative comparison of the ablation study on different inference steps.}
Step: Number of inference steps.}
\label{fig:App-InferStep}
\end{figure*}

We also examine the influence of the number of \textbf{inference steps}, as shown in Fig. \ref{fig:App-InferStep}. Increasing the inference steps leads to richer visual details in the generated images, at the cost of higher inference time.

Based on these observations, we adopt a balanced configuration for inference, setting the control strength to 1.0, the condition strength to 2.0, and the number of inference steps to 50, which achieves a favorable trade-off between image quality, realism, and computational efficiency.

\subsection{Additional Visual Results}
\begin{figure*}[t]
\centerline{\includegraphics[width=1.0\textwidth]{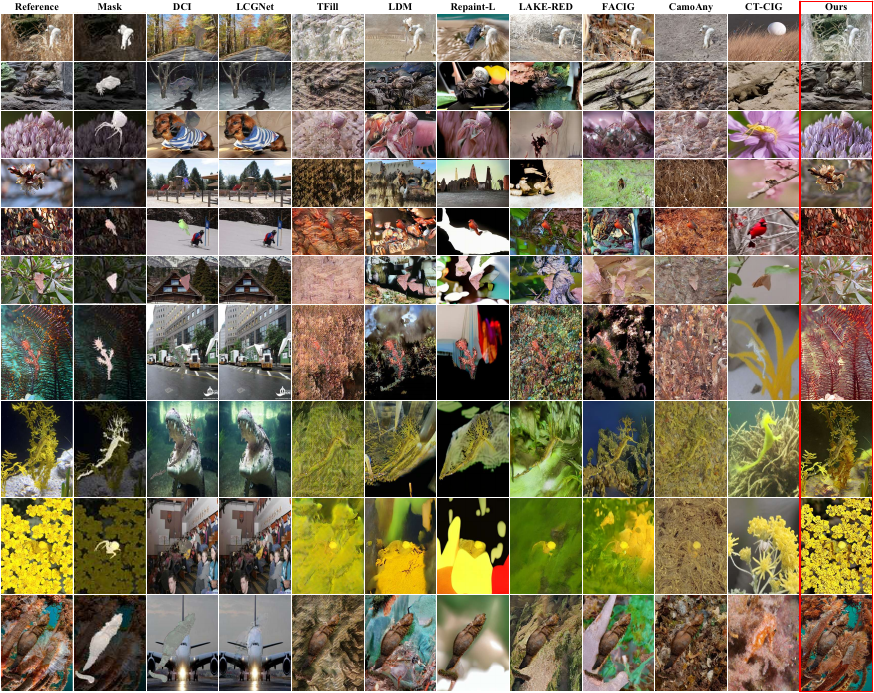}}
\caption{\textbf{Additional qualitative comparison with existing SOTA methods} on camouflaged objects transfer.
\textbf{Our results are demonstrated in the rightmost column with \textcolor{red}{red} box.}}
\label{fig:App-AdditionCOD}
\end{figure*}

\begin{figure*}[t]
\centerline{\includegraphics[width=1.0\textwidth]{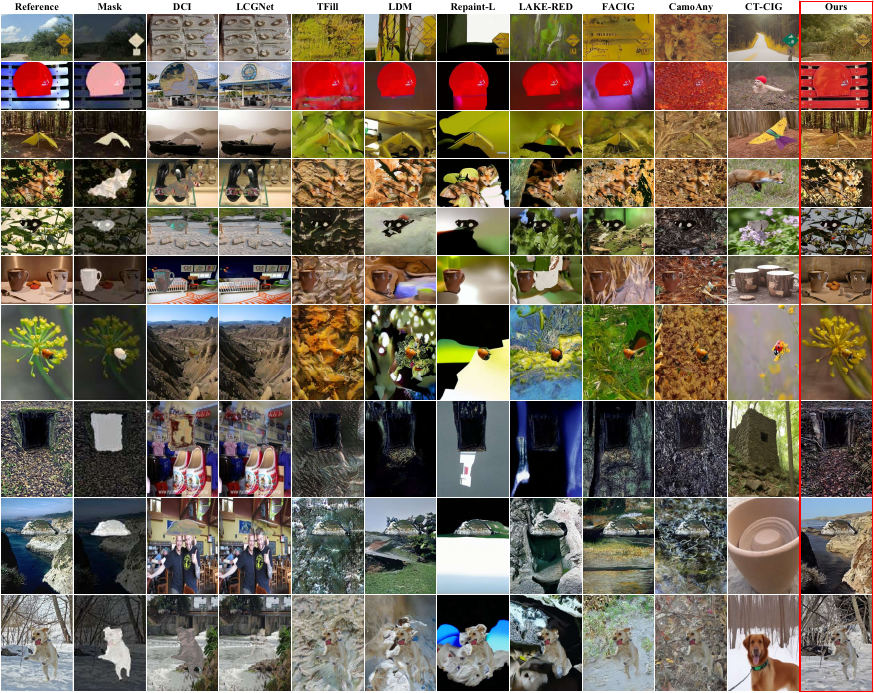}}
\caption{\textbf{Additional qualitative comparison with existing SOTA methods} on salient objects transfer.
\textbf{Our results are demonstrated in the rightmost column with \textcolor{red}{red} box.}}
\label{fig:App-AdditionSOD}
\end{figure*}

\begin{figure*}[t]
\centerline{\includegraphics[width=1.0\textwidth]{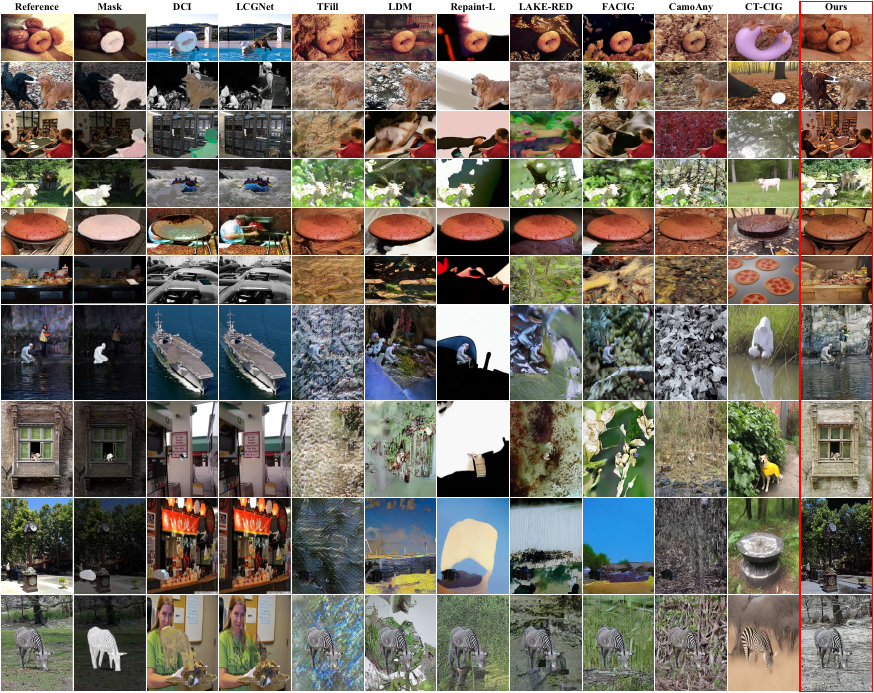}}
\caption{\textbf{Additional qualitative comparison with existing SOTA methods} on general objects transfer.
\textbf{Our results are demonstrated in the rightmost column with \textcolor{red}{red} box.}}
\label{fig:App-AdditionGOD}
\end{figure*}

To more comprehensively demonstrate the effectiveness of our RealCamo framework, we present additional qualitative results across three camouflage transfer scenarios, \textit{i.e.}, camouflaged objects transfer, salient objects transfer, and general objects transfer, as shown in Figs. \ref{fig:App-AdditionCOD}, \ref{fig:App-AdditionSOD}, and \ref{fig:App-AdditionGOD}. These results highlight the superior performance and robustness of our proposed method under diverse object characteristics and transfer settings.

\begin{table*}[htbp]
\begin{center}
\caption{\textbf{Performance of MVANet \cite{CVPR2024MVANet} on SOD and GOD with different training data.} The best results are highlighted in \textbf{bold}. $\uparrow$ / $\downarrow$ represents the higher / lower the better.}
\label{tab:App-SOD-GOD}
\renewcommand{\arraystretch}{1.2}
\setlength{\tabcolsep}{3pt}
\resizebox{1.0\textwidth}{!}{
\begin{tabular}{l|ccccccccc|ccccccccc}
\hline
\multirow{2}{*}{\textbf{Training Data}} &
\multicolumn{9}{c|}{\textbf{Salient Object Detection (1,000)}} &
\multicolumn{9}{c}{\textbf{General Object Detection (1,000)}} \\
\cline{2-19}
&
$S_\alpha \uparrow$&$E_\phi^m \uparrow$&$E_\phi^a \uparrow$&$E_\phi^x \uparrow$&$F_\beta^m \uparrow$&$F_\beta^a \uparrow$&$F_\beta^x \uparrow$&$F_\beta^w \uparrow$&$M \downarrow$&
$S_\alpha \uparrow$&$E_\phi^m \uparrow$&$E_\phi^a \uparrow$&$E_\phi^x \uparrow$&$F_\beta^m \uparrow$&$F_\beta^a \uparrow$&$F_\beta^x \uparrow$&$F_\beta^w \uparrow$&$M \downarrow$\\
\hline

Original &
\textbf{0.911} & \textbf{0.933} & \textbf{0.934} & \textbf{0.943} & \textbf{0.885} & \textbf{0.876} & \textbf{0.902} & \textbf{0.869} & \textbf{0.033} & 
\textbf{0.745} & \textbf{0.783} & \textbf{0.833} & \textbf{0.835} & \textbf{0.656} & \textbf{0.683} & \textbf{0.688} & \textbf{0.626} & \textbf{0.086} \\

Ori. + Gen $_{S / G}$ &
0.901 & 0.926 & 0.923 & 0.939 & 0.873 & 0.858 & 0.897 & 0.849 & 0.039 & 
0.730 & 0.767 & 0.815 & 0.816 & 0.635 & 0.675 & 0.670 & 0.606 & 0.095 \\
\hline

\end{tabular}}
\end{center}
\end{table*}

\begin{table*}[htbp]
\begin{center}
\caption{\textbf{Performance of MVANet \cite{CVPR2024MVANet} on the CHAMELEON and CAMO test set with different training data.} The best results are highlighted in \textbf{bold}. $\uparrow$ / $\downarrow$ represents the higher / lower the better.}
\label{tab:App-COD-1}
\renewcommand{\arraystretch}{1.2}
\setlength{\tabcolsep}{3pt}
\resizebox{1.0\textwidth}{!}{
\begin{tabular}{l|ccccccccc|ccccccccc}
\hline
\multirow{2}{*}{\textbf{Training Data}} &
\multicolumn{9}{c|}{\textbf{CHAMELEON (76)}} &
\multicolumn{9}{c}{\textbf{CAMO (250)}} \\
\cline{2-19}
&
$S_\alpha \uparrow$&$E_\phi^m \uparrow$&$E_\phi^a \uparrow$&$E_\phi^x \uparrow$&$F_\beta^m \uparrow$&$F_\beta^a \uparrow$&$F_\beta^x \uparrow$&$F_\beta^w \uparrow$&$M \downarrow$&
$S_\alpha \uparrow$&$E_\phi^m \uparrow$&$E_\phi^a \uparrow$&$E_\phi^x \uparrow$&$F_\beta^m \uparrow$&$F_\beta^a \uparrow$&$F_\beta^x \uparrow$&$F_\beta^w \uparrow$&$M \downarrow$\\
\hline

Original &
0.915 & 0.929 & \textbf{0.957} & 0.953 & 0.871 & 0.867 & 0.895 & 0.846 & 0.022 & 
0.870 & 0.896 & 0.918 & 0.927 & 0.842 & 0.847 & 0.865 & 0.809 & 0.055 \\

Ori. + Gen $_{S\&G}$ &
0.905 & 0.927 & 0.911 & 0.971 & 0.846 & 0.811 & 0.904 & 0.796 & 0.031 & 
0.826 & 0.842 & 0.881 & 0.885 & 0.760 & 0.784 & 0.795 & 0.717 & 0.079 \\

\rowcolor{gray!15}
Ori. + SynCOD12K &
\textbf{0.924} & \textbf{0.949} & 0.952 & \textbf{0.965} & \textbf{0.889} & \textbf{0.877} & \textbf{0.912} & \textbf{0.871} & \textbf{0.019} & 
\textbf{0.882} & \textbf{0.916} & \textbf{0.927} & \textbf{0.935} & \textbf{0.856} & \textbf{0.852} & \textbf{0.881} & \textbf{0.826} & \textbf{0.049} \\
\hline

\end{tabular}}
\end{center}
\end{table*}

\begin{table*}[htbp]
\begin{center}
\caption{\textbf{Performance of MVANet \cite{CVPR2024MVANet} on the COD10K and NC4K test set with different training data.} The best results are highlighted in \textbf{bold}. $\uparrow$ / $\downarrow$ represents the higher / lower the better.}
\label{tab:App-COD-2}
\renewcommand{\arraystretch}{1.2}
\setlength{\tabcolsep}{3pt}
\resizebox{1.0\textwidth}{!}{
\begin{tabular}{l|ccccccccc|ccccccccc}
\hline
\multirow{2}{*}{\textbf{Training Data}} &
\multicolumn{9}{c|}{\textbf{COD10K (2,026)}} &
\multicolumn{9}{c}{\textbf{NC4K (4,021)}} \\
\cline{2-19}
&
$S_\alpha \uparrow$&$E_\phi^m \uparrow$&$E_\phi^a \uparrow$&$E_\phi^x \uparrow$&$F_\beta^m \uparrow$&$F_\beta^a \uparrow$&$F_\beta^x \uparrow$&$F_\beta^w \uparrow$&$M \downarrow$&
$S_\alpha \uparrow$&$E_\phi^m \uparrow$&$E_\phi^a \uparrow$&$E_\phi^x \uparrow$&$F_\beta^m \uparrow$&$F_\beta^a \uparrow$&$F_\beta^x \uparrow$&$F_\beta^w \uparrow$&$M \downarrow$\\
\hline

Original &
0.882 & 0.917 & 0.913 & 0.944 & 0.828 & 0.792 & 0.854 & 0.792 & 0.025 &
0.886 & 0.910 & \textbf{0.919} & \textbf{0.939} & 0.855 & \textbf{0.844} & 0.879 & 0.820 & 0.038 \\

Ori. + Gen $_{S\&G}$ &
0.865 & 0.896 & 0.849 & 0.940 & 0.784 & 0.708 & 0.849 & 0.719 & 0.035 &
0.879 & 0.897 & 0.883 & 0.934 & 0.821 & 0.788 & 0.872 & 0.769 & 0.048 \\

\rowcolor{gray!15}
Ori. + SynCOD12K &
\textbf{0.896} & \textbf{0.934} & \textbf{0.919} & \textbf{0.953} & \textbf{0.850} & \textbf{0.805} & \textbf{0.877} & \textbf{0.818} & \textbf{0.021} &
\textbf{0.891} & \textbf{0.919} & 0.914 & 0.938 & \textbf{0.860} & 0.840 & \textbf{0.885} & \textbf{0.824} & \textbf{0.037} \\
\hline

\end{tabular}}
\end{center}
\end{table*}

\section{Downstream Applications}
\label{Sec:App-DownStreamApp}
\subsection{Model and Evaluation Metrics}
\textbf{Model.}
For downstream application evaluation, we adopt MVANet \cite{CVPR2024MVANet} as the segmentation model. MVANet is a dichotomous image segmentation (DIS) model designed for challenging fine-grained object segmentation, typically operating on high-resolution inputs (\textit{e.g.}, $1024 \times 1024$).
Since high-resolution segmentation accuracy is not the primary focus of this work and the native output resolution of our RealCamo framework is $512 \times 512$, we conduct all training, inference, and evaluation at a unified resolution of $512 \times 512$ using MVANet.

\textbf{Evaluation Metrics.}
To enable a more comprehensive evaluation, we adopt all the following metrics:
mean absolute error ($M$), E-measures (including mean E-measure ($E_\phi^m$), average E-measure ($E_\phi^a$), and max E-measure ($E_\phi^x$)), F-measures (including mean F-measure ($F_{\beta}^m$), average F-measure ($F_{\beta}^a$), and max F-measure ($F_{\beta}^x$)), weighted F-measure ($F_{\beta}^{w}$), and structure measure ($S_{\alpha}$). Better performance is indicated by lower values of $M$ and higher values of the others.

\subsection{Experiments on SOD and GOD}
For the SOD task (and analogously for the GOD task), we randomly select 1,000 images from the corresponding LAKE-RED validation subset as the test set, with the remaining 5,473 images used for training. We conduct two experimental settings: \textbf{(i)} training the model using the original images only (denoted as ``Original''), and \textbf{(ii)} training with both the original images and their camouflage-transferred counterparts (denoted as ``Ori. + Gen $_{S / G}$''). The quantitative results are reported in Tab. \ref{tab:App-SOD-GOD}.

The results show that incorporating camouflage-transferred images as additional training data significantly degrades performance on both SOD and GOD tasks. This outcome is expected, as camouflage transfer inherently suppresses saliency cues and blurs category-specific visual characteristics, thereby obscuring the discriminative signals required for accurate detection. Consequently, models trained with such data struggle to correctly localize salient or category-specific targets, leading to inferior performance.

\begin{figure*}[htbp]
\centerline{\includegraphics[width=0.75\textwidth]{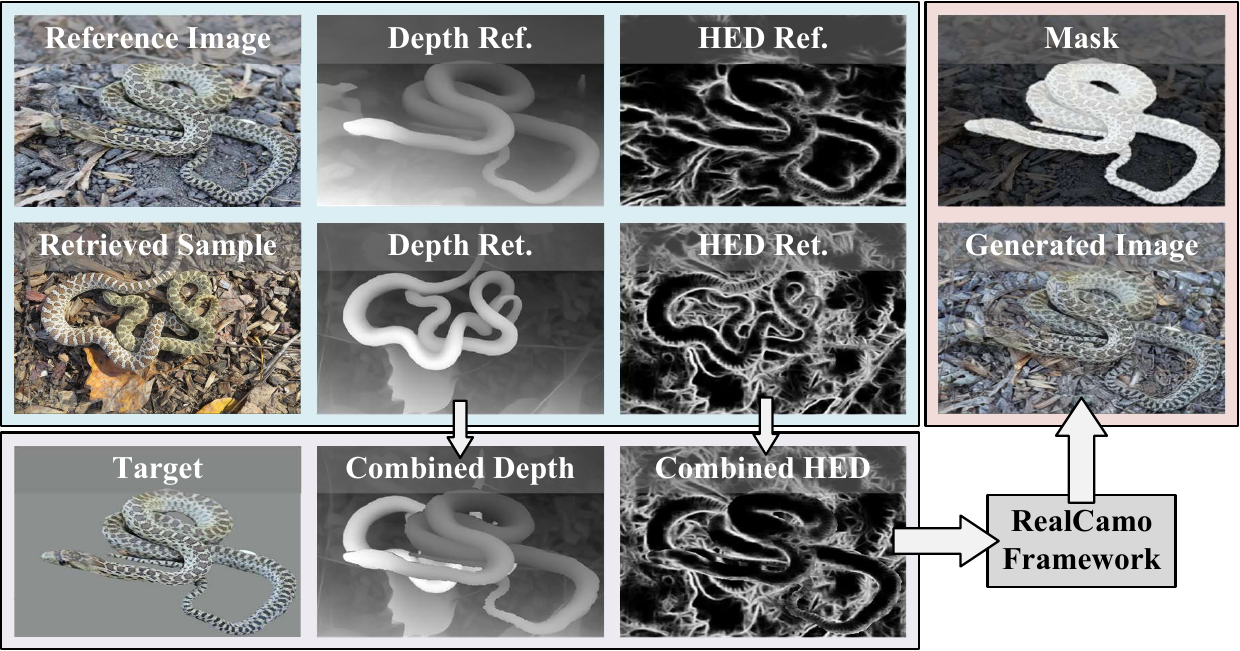}}
\caption{\textbf{Pipeline of constructing synthetic camouflaged images in the SynCOD12K dataset.}}
\label{fig:App-SynCOD}
\end{figure*}

\subsection{Experiments on COD}
\textbf{The SynCOD12K Dataset.}
To enrich training data for COD, we construct a synthetic dataset termed \textbf{SynCOD12K} using the proposed RealCamo framework. To avoid data leakage, we start from the standard COD training split consisting of 4,040 images. For each image, the Texture-Oriented Background Retrieval (TOBR) module is employed to retrieve the top-3 background reference samples.
Specifically, given an input image $\mathrm{I}$ and its retrieved reference samples $\{ S_j \}_{j=1}^3$, we first extract the corresponding depth maps and HED edge maps, which are then unified to the same resolution and fused to produce a combined depth map and a combined HED map. These structural cues, together with the textual–visual conditions, are fed into the RealCamo pipeline to generate high-quality camouflaged images. The overall data generation process is illustrated in Fig. \ref{fig:App-SynCOD}.

\textbf{Evaluation Results.}
We evaluate downstream COD performance under three training configurations: \textbf{(i)} using the standard COD training set (denoted as ``Original''), \textbf{(ii)} augmenting (i) with camouflaged images transferred from the salient and general subsets of LAKE-RED validation set (denoted as ``Ori. + Gen $_{S\&G}$''), and \textbf{(iii)} augmenting (i) with the proposed SynCOD12K dataset (denoted as ``Ori. + SynCOD12K'').

Results in Tabs. \ref{tab:App-COD-1} and \ref{tab:App-COD-2} show that salient- and general-synthesis camouflaged images degrade COD performance, likely due to a remaining domain gap despite camouflage transfer. In contrast, incorporating SynCOD12K consistently improves performance, indicating that realistically synthesized camouflaged images can effectively benefit downstream COD training.

\section{User Study}
\label{Appendix:UserStudy}
To further evaluate the quality of generated camouflaged images, we design a more comprehensive human-feedback-based evaluation framework and conduct a user study. We prepare two parallel questionnaires (\textit{i.e.}, \textit{Version A} and \textit{Version B}), each consisting of two sections. A total of 20 participants were recruited, with 10 participants assigned to each version.

We thank \textbf{\textit{Wenchao Liu}} and \textbf{\textit{Hengrui Qu}} for their contributions to the questionnaire design, and we also sincerely express our appreciation to the 20 participants for providing valuable human feedback results.

\begin{figure*}[t]
\centerline{\includegraphics[width=0.8\textwidth]{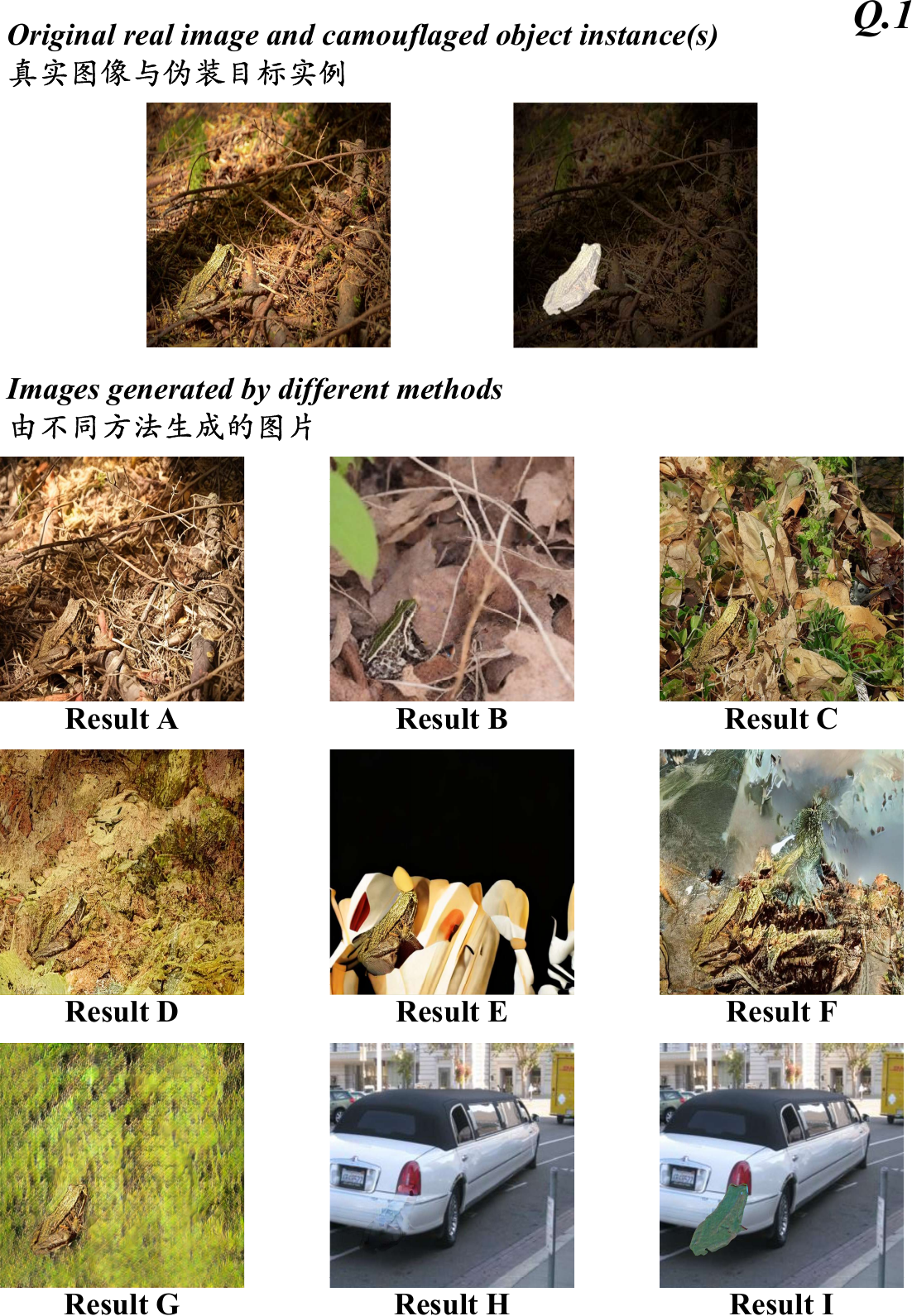}}
\caption{\textbf{An example} for Part I of the user study (\textit{Version A}).}
\label{fig:UserStudy1}
\end{figure*}

\begin{figure*}[t]
\centerline{\includegraphics[width=0.8\textwidth]{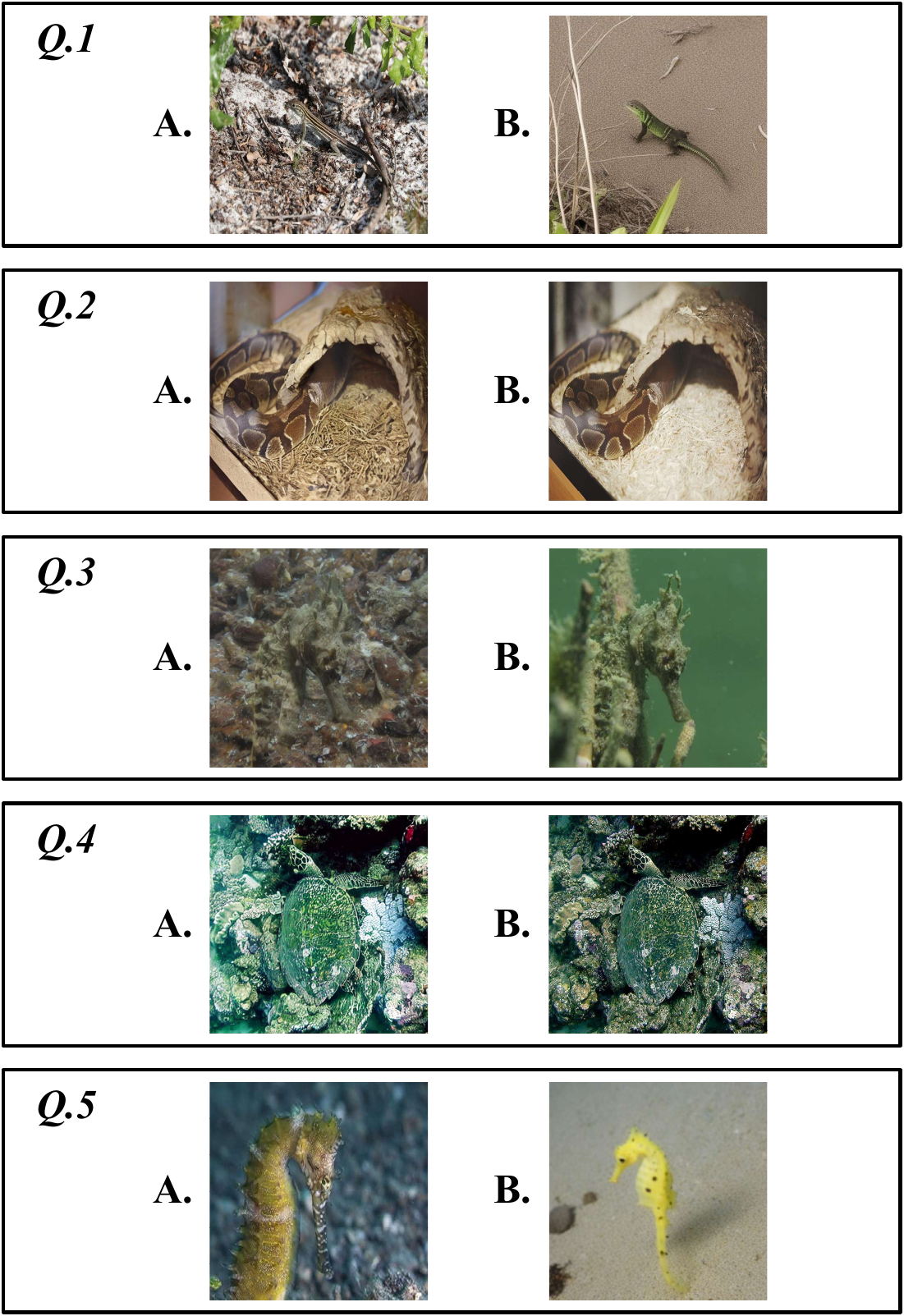}}
\caption{\textbf{Some examples} for Part II of the user study (\textit{Version A}).}
\label{fig:UserStudy2}
\end{figure*}

The questionnaires will be submitted as supplementary material, with some examples illustrated in Figs. \ref{fig:UserStudy1} and \ref{fig:UserStudy2}.

\begin{figure*}[htbp]
\centerline{\includegraphics[width=1.0\textwidth]{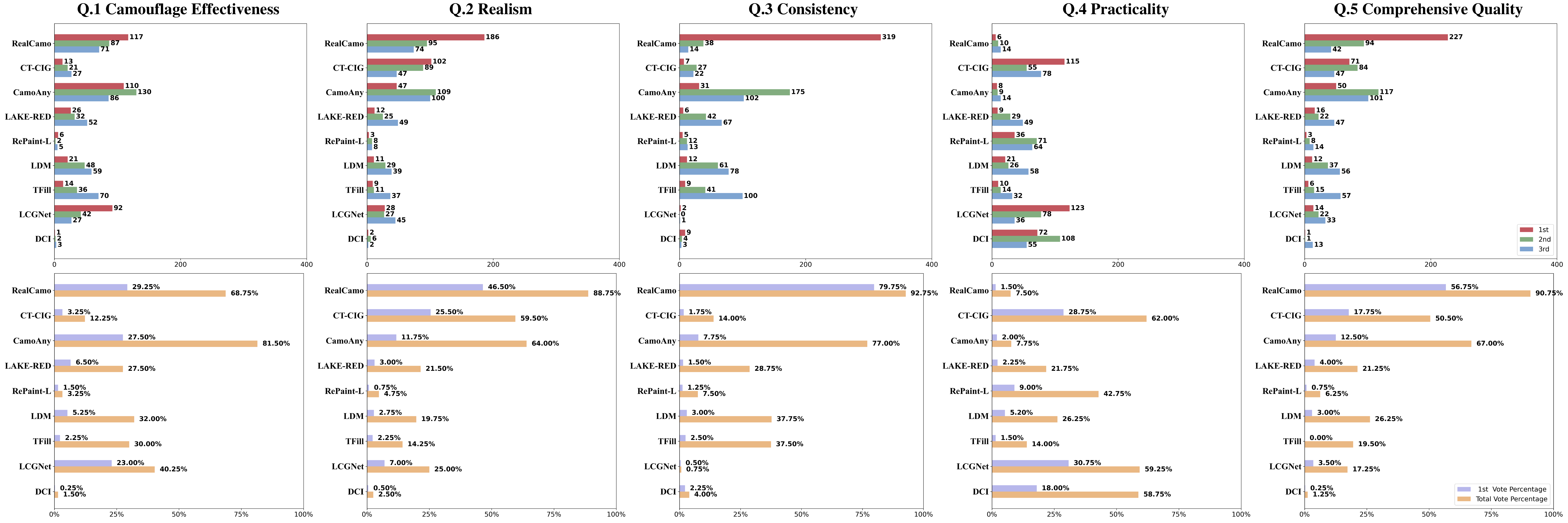}}
\caption{\textbf{Human feedback results for Part I of the user study.}
\textbf{Top:} the \textbf{\textcolor{1st}{1st}}, \textbf{\textcolor{2nd}{2nd}}, and \textbf{\textcolor{3rd}{3rd}} vote numbers on five questions.
\textbf{Bottom:} the \textbf{\textcolor{top1p}{1st}} and \textbf{\textcolor{totalp}{total}} vote percentages on five questions.
Please zoom in and view in color for more details.
}
\label{fig:UserStudy}
\end{figure*}

\subsection{Part I: Multi-Dimensional Evaluation}
In this part, we conduct a comprehensive multi-dimensional evaluation of camouflaged image generation, covering nine different methods inclueing DCI \cite{AAAI2020DCI}, LCGNet \cite{TMM2022LCGNet}, TFill \cite{CVPR2022TFill}, LDM \cite{CVPR2022LDM}, RePaint-L \cite{CVPR2022Repaint}, LAKE-RED \cite{CVPR2024LAKERED}, CamoAny \cite{CVPR2025CamoAny}, CT-CIG \cite{AAAI2026CTCIG}, and our proposed RealCamo.

This section contains 20 samples. For each sample, we provide a real camouflaged image (captured in nature) and the camouflaged targets contained in this image. We also present synthetic results generated by nine different methods. For each sample, participants are required to answer the following five questions. For each question, they should provide the \textbf{top-3 most suitable answers in order of priority}.

\begin{itemize}
    \item \textbf{Q.1 Camouflage Effectiveness:} Which result makes the camouflaged target hardest to detect? \textit{Consider both detection difficulty and validity of camouflage.}
    \item \textbf{Q.2 Realism:} Which result looks most like a real image? \textit{Only evaluate the nine displayed results, excluding the reference real image.}
    \item \textbf{Q.3 Consistency:} Which result is most similar to the reference real image? \textit{Consider image structure and the semantics in the content.}
    \item \textbf{Q.4 Practicality:} Which result has \textbf{the largest deviation} between the camouflaged target area and the ground truth? \textit{i.e., the mismatch degree between the target position in the generated result and the actual position.}
    \item \textbf{Q.5 Comprehensive Quality:} Which result has the highest image quality? \textit{Evaluate solely from the perspective of a normal image/photo.}
\end{itemize}

To ensure the impartiality of the user study, the display order of the results from the nine methods is fully randomized for every example. In addition, \textit{Version A} follows the sample selection protocol used in the LAKE-RED user study, whereas \textit{Version B} consists of randomly selected samples.

Fig. \ref{fig:UserStudy} summarizes the human feedback results across the five assessment dimensions: camouflage effectiveness, realism, consistency, practicality, and comprehensive quality. RealCamo consistently achieves the highest human preference scores on all five dimensions, indicating its superior ability to generate visually coherent and convincingly camouflaged images. Notably, the most recent approaches, CamoAny and CT-CIG, also exhibit competitive performance, particularly in camouflage effectiveness and realism, suggesting that recent advances in camouflaged image generation have substantially improved perceptual fidelity.

\begin{table*}[htbp]
\begin{center}
\caption{\textbf{Human feedback results for Part II of the user study.} The highest values are highlighted in \textbf{bold}.
A lower accuracy means a higher misjudgment rate, indicating generated images with more realism.}
\label{tab:UserStudy}
\renewcommand{\arraystretch}{1.2}
\setlength{\tabcolsep}{3pt}
\resizebox{0.75\textwidth}{!}{
\begin{tabular}{l|c|cccc}
\hline
\multirow{2}{*}{\textbf{Method}} &
\multirow{2}{*}{\textbf{Q.1 Accuracy}} &
\multicolumn{4}{c}{\textbf{Q.2 Confidence}} \\
\cline{3-6}
 & & Option.A & Option.B & Option.C & Option.D \\
\hline

CamoAny &
\textbf{79.50\%} & \textbf{51.00\%} & 23.50\% & 14.00\% & 11.50\% \\

CT-CIG &
66.00\% & 49.50\% & \textbf{26.50\%} & 9.00\% & 15.00\% \\

\rowcolor{gray!15}
RealCamo &
43.00\% & 18.50\% & 20.50\% & \textbf{22.50\%} & \textbf{38.50\%} \\
\hline

\end{tabular}}
\end{center}
\end{table*}

\subsection{Part II: Further Realism Evaluation}
In this part, we further evaluate the realism of the generated camouflaged images.

This section contains 30 samples. For methods with probably top-3 realism in Part I (\textit{i.e.}, RealCamo, CT-CIG, and CamoAny), we randomly select 10 samples for each. For each sample, we provide a real camouflaged image (captured in nature) and a camouflaged image generated by one of the abovementioned three methods. For each sample, participants are required to answer the following two questions. For each question, they should provide \textbf{the most suitable answer}.

\begin{itemize}
    \item \textbf{Q.1:} Which result is the real camouflaged image? \textit{i.e., the scene should naturally exist in the real world.}
    \item \textbf{Q.2:} What is your experience in making the judgment, and how confident are you in your choice? \textit{Please select one option from the following.}
    \begin{itemize}
        \item \textbf{Option.A:} I can instantly distinguish the real image and am \textbf{fully certain} of my judgment.
        \item \textbf{Option.B:} I need a short while to make the judgment, but I am \textbf{certain} of my judgment.
        \item \textbf{Option.C:} I have to deliberate carefully to make the judgment, but I am \textbf{certain} of my judgment.
        \item \textbf{Option.D:} I have to deliberate carefully to make the judgment and still \textbf{remain uncertain} about my judgment.
    \end{itemize}
\end{itemize}

To ensure the impartiality of the user study, for each sample, the method used for generating the result, and the display order of the real image are completely random.

Tab. \ref{tab:UserStudy} reports human feedback results on further realism evaluation, including the judgment accuracy and confidence. RealCamo induces the highest misjudgment rate (\textit{i.e.}, the lowest accuracy), requires a longer average decision time, and elicits the greatest uncertainty among participants. These results demonstrate that images produced by RealCamo are perceptually challenging even for human observers, which further validates the effectiveness of our proposed framework.

\section{Limitations and Future Work}
Despite the superior performance of RealCamo in realistic camouflaged image synthesis and its demonstrated benefits for downstream COD tasks, several limitations remain and motivate future research.

\begin{figure*}[htbp]
\centerline{\includegraphics[width=1.0\textwidth]{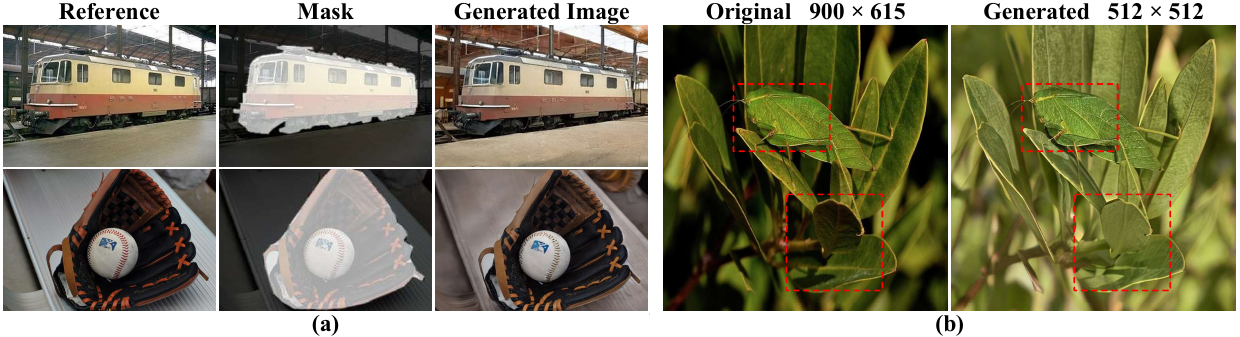}}
\caption{\textbf{Limitations of the proposed RealCamo framework.}
\textbf{(a)} Camouflage transfer from salient and general images remains challenging.
\textbf{(b)} Compared to high-resolution real camouflaged images, the generated lower-resolution result lacks fine details.} 
\label{fig:App-Limitations}
\end{figure*}

First, \textbf{camouflage transfer from salient and general images} remains challenging, as shown in Fig. \ref{fig:App-Limitations} (a). Although RealCamo achieves reasonable transfer results, the synthesized images transferred from salient and general images still exhibit weaker visual camouflage. This may stem from both the domain gap between naturally camouflaged training data and non-camouflaged source images, as well as the interaction between explicit structural controls and pretrained diffusion priors, which can impede the emergence of strong camouflage cues. Developing mechanisms that enhance visual camouflage while preserving structural consistency is an important direction.

Second, \textbf{generation diversity} is inherently constrained by structural controls. While such controls improve realism and semantic consistency, they limit variability in the generated scenes. Although our controlled information fusion (as shown in Fig. \ref{fig:App-SynCOD}) partially mitigates this issue, designing principled methods that jointly support structural guidance and diverse generation remains an open problem.

Third, \textbf{image resolution and fine-grained details} are limited by the underlying Stable Diffusion 1.5 backbone, which operates at a $512 \times 512$ resolution. Compared to high-resolution real camouflaged images, the generated results lack fine details, as demonstrated in Fig. \ref{fig:App-Limitations} (b). Leveraging more advanced high-resolution generative models (\textit{e.g.}, SDXL \cite{ICLR2024SDXL} or FLUX \cite{arXiv2025FLUX}) is a promising avenue for improving visual fidelity.

Finally, while we show that synthesized camouflaged images can benefit downstream COD training, \textbf{how to more effectively integrate synthetic data} into COD remains underexplored and warrants further investigation, with some pioneer work \cite{ACMMM2025SynCOD, arXiv2026GenCAMO} here for reference.

\newpage

\end{document}